\def\year{2022}\relax
\documentclass[letterpaper]{article} 
\usepackage{aaai22}  
\usepackage{times}  
\usepackage{helvet}  
\usepackage{courier}  
\usepackage[hyphens]{url}  
\usepackage{graphicx} 
\usepackage{natbib}  
\usepackage{caption} 
\DeclareCaptionStyle{ruled}{labelfont=normalfont,labelsep=colon,strut=off} 
\usepackage{algorithm}
\usepackage{algorithmic}

\usepackage{newfloat}
\usepackage{listings}

\usepackage{caption,subcaption}
\usepackage{amsmath,amsthm}
\usepackage{colortbl}
\usepackage{arydshln}
\usepackage{multirow}
\usepackage{multicol}
\usepackage{graphicx} 
\usepackage{mdwlist}
\usepackage{caption,subcaption}
\usepackage{amsmath,amsthm,amssymb}
\DeclareMathOperator*{\argmax}{arg\,max}

\floatstyle{ruled}
\newfloat{listing}{tb}{lst}{}
\floatname{listing}{Listing}
\title{An Entropy-guided Reinforced Partial Convolutional Network for Zero-Shot Learning}

\author {
    Yun Li,\textsuperscript{\rm 1}
    Zhe Liu, \textsuperscript{\rm 1}
    Lina Yao, \textsuperscript{\rm 1}
    Xianzhi Wang, \textsuperscript{\rm 2}
    Julian McAuley, \textsuperscript{\rm 3}
    Xiaojun Chang, \textsuperscript{\rm 4}
}
\affiliations {
    \textsuperscript{\rm 1} University of New South Wales\\
    \textsuperscript{\rm 2} AUniversity of Technology Sydney\\
    \textsuperscript{\rm 3} University of California San Diego \\
    \textsuperscript{\rm 4} RMIT University\\
    yun.li5@unsw.edu.au, she.liu1@student.unsw.edu.au, lina.yao@unsw.edu.au,
    XIANZHI.WANG@uts.edu.au,
    jmcauley@eng.ucsd.edu,
    cxj273@gmail.com
}

\usepackage{bibentry}

\begin{document}

\maketitle

\begin{abstract}
Zero-Shot Learning (ZSL) aims to transfer learned knowledge from observed classes to unseen classes via semantic correlations. A promising strategy is to learn a global-local representation that incorporates global information with extra localities (i.e., small parts/regions of inputs). However, existing methods discover localities based on explicit features without digging into the inherent properties and relationships among regions. In this work, we propose a novel Entropy-guided Reinforced Partial Convolutional Network (ERPCNet), which extracts and aggregates localities progressively based on semantic relevance and visual correlations without human-annotated regions. 
ERPCNet uses reinforced partial convolution and entropy guidance; it not only discovers global-cooperative localities dynamically but also converges faster for policy gradient optimization. We conduct extensive experiments to demonstrate ERPCNet's performance through comparisons with state-of-the-art methods under ZSL and Generalized Zero-Shot Learning (GZSL) settings on four benchmark datasets. 
We also show ERPCNet is time efficient and explainable through visualization analysis.
\end{abstract}

\section{Introduction}

Zero-shot Learning (ZSL) mimics the human ability to perceive unseen concepts. In image classification, a ZSL model should still work when only semantic descriptions (attributes that describe the visual characteristics of an image, e.g.~\textit{is black}) of a class are given. A typical scheme for ZSL is to extract visual representations from images and then learn visual-semantic associations. However, these approaches focus on global features while failing to capture subtle local differences between classes. A few works have paved the way to incorporate `locality' knowledge, i.e., discriminative parts/regions in the original image, into global information~\cite{zhu2019semantic,xu2020attribute,sylvain2019locality,hjelm2018learning}. These approaches can be annotation-based or weakly-supervised~\cite{zhu2019semantic,xu2020attribute,sylvain2019locality,ji2018stacked,xie2019attentive}. Annotation-based methods~\cite{akata2016multi,elhoseiny2017link,ji2018stacked} use extra annotations of important local regions to supervise the locality learning.
Manual annotations are often time-consuming and costly to obtain.
Weakly-supervised methods mitigate this by detecting
salient local regions without ground-truth annotations.
They adopt multi-attention~\cite{zhu2019semantic,huynh2020fine,xie2019attentive,huynh2020compositional} or pre-defined strategies~\cite{xu2020attribute,sylvain2019locality} to capture diverse localities. 

\begin{figure}
    \centering
    \begin{subfigure}{0.21\textwidth}
    \centering
    \includegraphics[width=\textwidth]{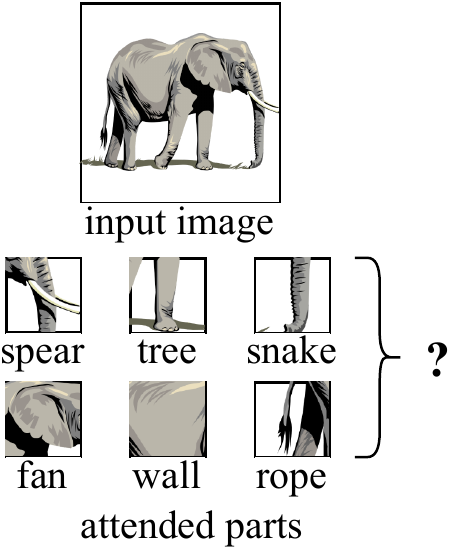}
    \caption{Conventional locality.}
    \end{subfigure}\hfil 
    \begin{subfigure}{0.21\textwidth}
    \includegraphics[width=\linewidth]{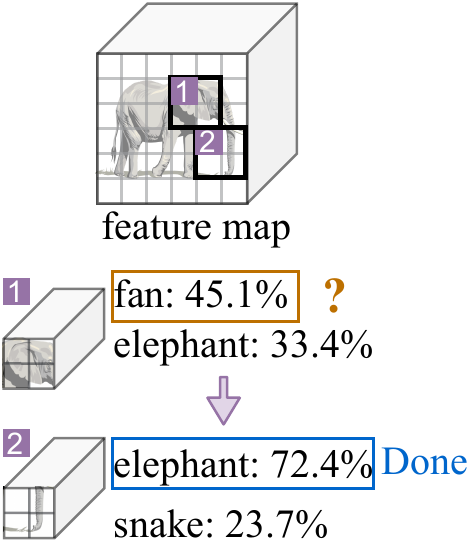}
    \caption{Progressive locality.}
    \end{subfigure}\hfil 
\caption{Locality comparisons.}
\label{intro}
\end{figure}

\begin{figure*}[h]
\centering
  \includegraphics[width=0.8\linewidth]{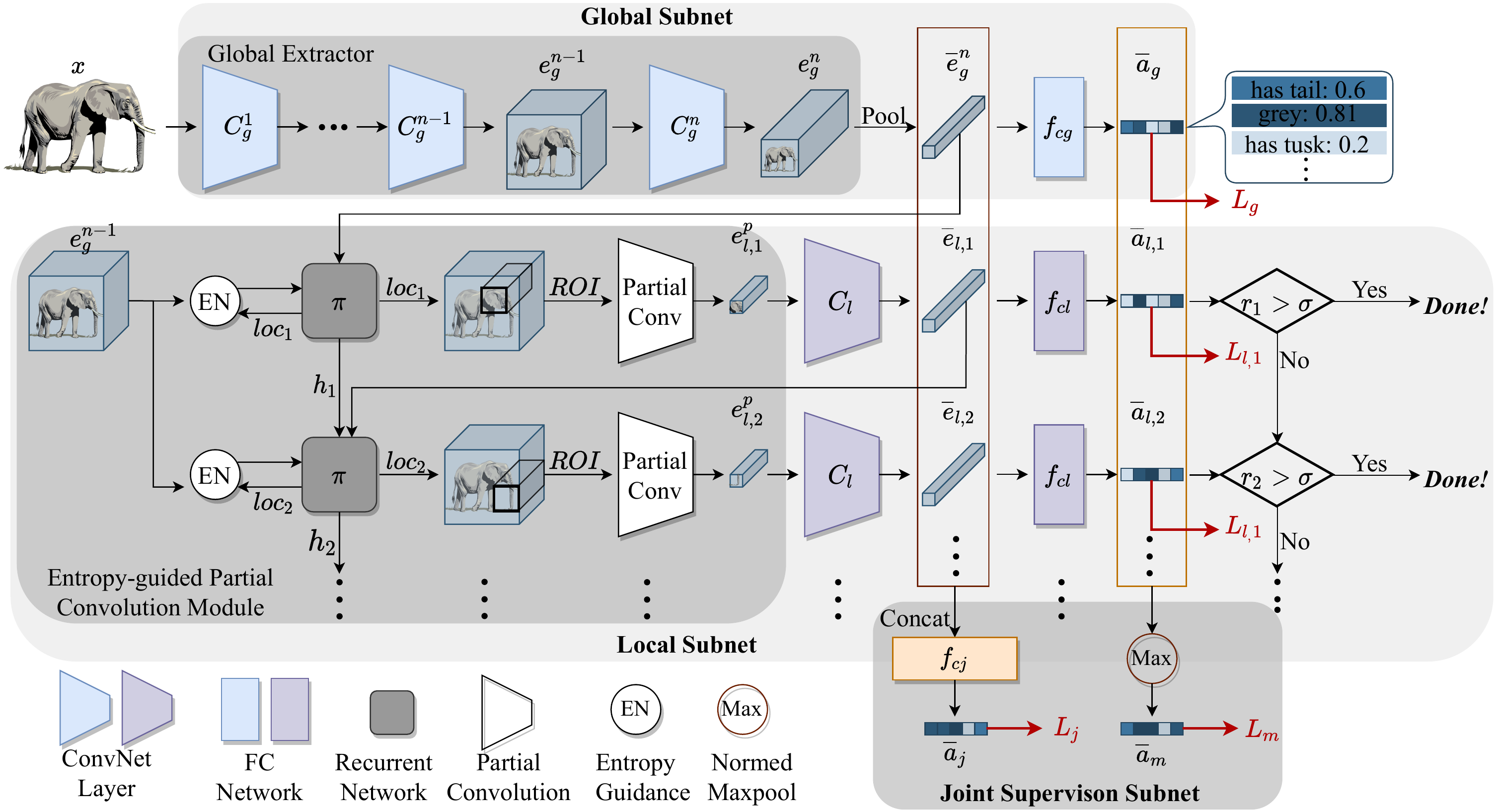}
  \caption{Overview of ERPCNet. Given an input image $x$, the model extracts global embeddings $\overline{e}_g^n$ from $x$ and progressively processes a sequence of local regions at the abstraction hierarchies located at $\{\mathit{loc}_1,\mathit{loc}_2,\dots\}$. At $t$\mbox{-}th step, ERPCNet conducts partial convolution and local extraction on the local region, as well as selecting the next location using an entropy-guided sampler $\pi$. The global and local visual features are fed into the corresponding predictors $f_{cg}$ and $f_{cl}$, respectively, for zero-shot recognition.
  The local loss $L_l$ guarantees the distinctiveness of a locality and its embedding $\overline{e}_{l,t}$.
  The joint supervision subnet optimizes the model to improve global-local cooperation ($L_j$) and strengthens 
  divergence across localities ($L_m$). Locality selection terminates once sufficient reward is obtained.
  }
  \label{method}
\end{figure*}

However, existing studies~\cite{zhu2019semantic,xu2020attribute,xie2019attentive} only consider fixed numbers of localities while neglecting that different images may need different numbers of localities. 
The need for
locality exploration increases when images are harder to classify (and decreases otherwise).
Using a fixed number of
localities 
can thus be inefficient and may introduce noise.
Moreover, these methods~\cite{huynh2020fine,zhu2019semantic,xie2019attentive} learn regions independently without accounting for inter-dependencies among regions, which
may lead to poor performance on downstream tasks.
For example, Figure~\ref{intro}(a) shows a conventional deep learning version of the blind man and the elephant parable. In this example, six attention-maps/extractors each extract a different part as the locality and tend to identify the elephant as different objects (namely snake, spear, etc.). Then, all the extracted localities will confuse the final classifier that aims to distinguish the elephant.

To address the above problems, we introduce Reinforcement Learning (RL) to progressively highlight localities based on region correlations. However, it is challenging to train the reinforced model under weak supervision and to scale to real-world datasets~\cite{sermanet2014attention}. Thus, we learn localities at the 
level of
abstraction hierarchies, i.e., convolution-level, to enable fast training. As shown in Figure~\ref{intro}(b), our model first selects an ear-related feature map and speculates the object as a fan, an elephant, etc, and then chooses the nose-related feature map based on former selections and recognizes the object as an elephant.

In this work, we propose a novel Entropy-guided Reinforced Partial Convolutional Network (ERPCNet) for effective global-local learning.
It leverages RL to learn localities progressively based on semantic relevance and inherent relationships among regions, which can improve the performance but explore fewer localities.
We design partial convolution to ease the sample efficiency problem for better RL optimization. 
Sample efficiency refers to the action amount needed for an RL agent to reach certain levels of performance. 
Partial convolution can reduce the action space by integrating RL in the abstraction hierarchies instead of at the conventional image-level to allow fast training.
It is also more efficient than processing image-level localities from scratch at each step. 
Besides, the entropy, introduced as expert knowledge, can complement the reward of the reinforced module to further accelerate reward learning. 

In summary, we make the following contributions:
\newline---We present ERPCNet for ZSL. ERPCNet learns global-local representations in a weakly-supervised manner. It adopts RL to progressively find localities 
that complement the
global representation based on semantic relevance and local relationships. The RL agent improves model performance with fewer locality proposals. We propose partial convolution to extract localities, which can mitigate high training cost of RL and reduce computation cost of extraction.
\newline---We design an entropy-guided reward function and use an entropy ratio to reflect the informativeness of localities.
The ratio can guide the training of the RL agent to converge faster. It also slightly improves the model's performance.
\newline---We carry out extensive experiments on four benchmark datasets in both ZSL and Generalized Zero-shot Learning (GZSL) settings to prove the improvement of our model over the state-of-the-art.
We further analyze and shed light on the effectiveness, efficiency, and explainability of our model. 

\section{Methodology}

We start by introducing the problem definition of ZSL/GZSL and notations used in the paper. 
Let $\mathcal{S}=\{(x, y, a)|x\in{X^{S}},y\in{Y^{S}},a\in{A^{S}}\}$ be the training data from seen classes (i.e.,~classes with labeled samples), where $x\in{X^{S}}$ denotes the data instance (i.e.,~an image), $y\in{Y^{S}}$ denotes the class label of $x$, and $a\in{A^{S}}$ represents an attribute (or other semantic side information) of $y$. 
Similarly, we define test data from unseen classes as $\mathcal{U}=\{(x, y, a)|x\in{X^{U}},y\in{Y^{U}},a\in{A^{U}}\}$.
Given an image $x$ from an unseen class and a set of attributes of unseen classes $A^{U}$, ZSL aims to predict the class label $y\in{Y^{U}}$ of the image, where seen and unseen classes are disjoint, i.e.,~$Y^{S}\cap{Y^{U}}= \emptyset$. 
GZSL is more challenging, aiming to predict images from both seen and unseen classes, i.e.,~$y\in Y^{U}\cup Y^{S}$.

\subsection{Overview}
The procedure of ERPCNet is described in Figure~\ref{method}.
ERPCNet consists of the global subnet, the local subnet, and the joint supervision subnet. The global subnet extracts global information and provides inspiration for determining the initial patch location. The local subnet adopts the entropy-guided sampler $\pi$ to select discriminative parts and then conducts partial convolution for locality extraction. The joint supervision subnet, composed of two branches, takes the global/local visual and semantic embeddings as the input to conduct joint supervision for better optimization.

The global subnet consists of the global extractor $f_{G}$ and the corresponding predictor $f_{cg}$. Let $e_{g}^{i}$ be the corresponding output of the $i$-th layer of $f_{G}$. The global extractor $f_{G}$ takes raw images as input and plays two important roles in the network: 1) extracting the global representation $\bar{e}_{g}^{n}$ of the original images and 2) providing the preliminary information $e_{g}^{n-1}$ for the local subnet, where $f_{cg}$ optimizes $\bar{e}_{g}^{n}$ and $e_{g}^{n-1}$ to carry the attribute information. 

Given $\bar{e}_{g}^{n}$ and $e_{g}^{n-1}$ produced by the global subnet, the local subnet employs the partial convolution module $f_{P}$, the locality extractor $C_{l}$ and the predictor $f_{cl}$ to progressively learn localities to complement our global representation. $f_{P}$ provides localities by an entropy-guided sampler $\pi$ (for region selection) and a convolution kernel (for partial convolution). $C_{l}$ further extracts high-level locality representation, and $f_{cl}$ ensures attribute-richness of the locality.

With the global representation and extracted localities from global/local subnets, the joint supervision subnet optimizes the extracted embeddings. It consists of a fusion module $f_{cj}$ and a normalized max pool for joint attribute regularization and highlighted attribute regularization, respectively. 

\subsection{Global Subnet}

The global subnet aims to extract discriminative global features for ZSL and provide adequate preliminary information for the local subnet. Given an input image $x$, the global extractor $f_G=\{C^{1}_{g},C^{2}_{g},...,C^{n}_{g}\}$ (a CNN backbone) embeds the input to a visual feature map $e^n_g \in \mathbb{R}^{H\times W \times \mathit{CH}}$: $e_{g}^{n} = C^{n}_{g}(...(C_{g}^{1}(x)))$, where $H$, $W$ and $\mathit{CH}$ denote height, width and channel, respectively; $n$ denotes the $n$\mbox{-}th layer in the global extractor; $C$ denotes a convolutional block.

The extractor is followed by global average pooling to learn a visual embedding $\bar{e}^n_g$, which is further projected into the semantic space by the predictor $f_{cg}$. $f_{cg}$ optimizes the global subnet using the loss $L_{g}$ to promote the compatibility between the embedding and the corresponding attribute:
\begin{equation}
    L_{g}  = \mathit{CE}(\bar{a}_{g},y)
     = -\log\frac{\exp(f_{cg}(\bar{e}_{g}^{n})^{T}\phi(y))}{\sum_{\hat{y}\in Y^{S}}\exp(f_{cg}(\bar{e}_{g}^{n})^{T}\phi(\hat{y}))}
\end{equation}
where $\bar{e}^{n}_{g}=\mathit{AdaptiveAvgPool}(e_{g}^{n})$; $y$ denotes the label for $x$; $\phi(y)$ denotes the attribute of $y$; $\mathit{CE}$ denotes CrossEntropy.

\begin{figure}
\centering
    \includegraphics[width=0.4\textwidth]{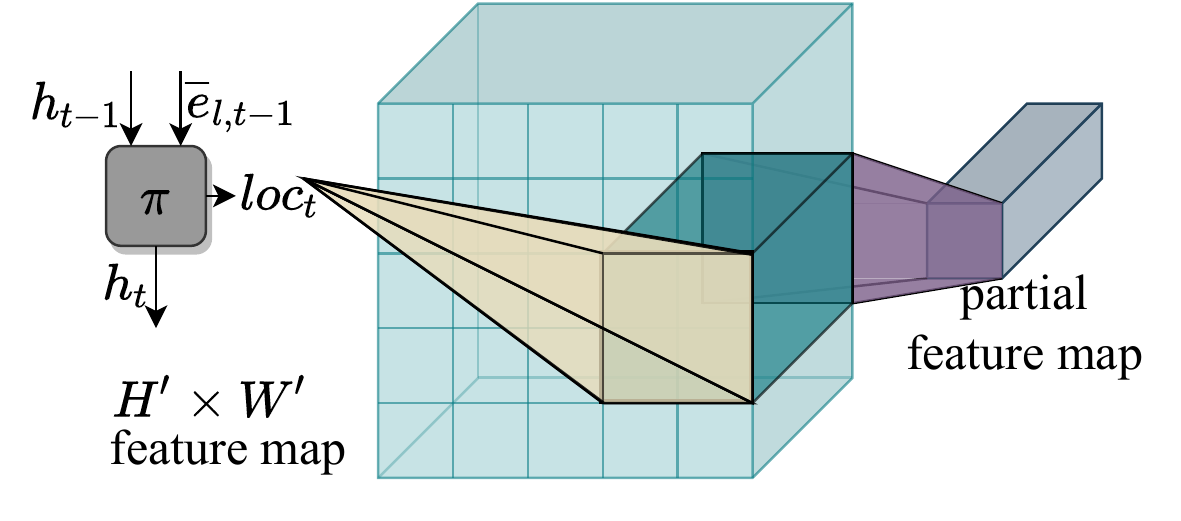}
  \caption{Partial convolution $f_P$.}
  \label{PartialConv}
\end{figure}

\subsection{Local Subnet}

The local subnet aims to progressively discover the localities $\bar{e}_{l,t}$ to complement to $\bar{e}^{n}_{g}$. We propose the entropy-guided reinforced partial convolution module $f_P$, the local extractor $C_l$, and the local predictor $f_{cl}$ to select regions and extract localities iteratively.

\textbf{Entropy-guided reinforced partial convolution.} Traditional convolution starts with a kernel that slides over the input data. The kernel repeatedly conducts element-wise multiplication and aggregates the results on all locations that it slides over. Unlike this, to explore and strengthen localities, we conduct partial convolution, i.e., we carry out the multiplication and summation procedure only on selected regions that are critical for classification, as shown in Figure~\ref{PartialConv}.

Suppose that $H'\times W'$ is the input size of the partial convolution, and $k \times k$, $q$, and $p$ are the kernel size, stride, and padding, respectively.
Partial region location search can be transferred to a grid search problem with the grid size of $(\frac{H'+2p-k}{q}+1)\times (\frac{W'+2p-k}{q}+1)$. 
The search is fulfilled by a recurrent network $\pi$ that aggregates all the previous information. 
Since partial convolution is non-differentiable, we consider $\pi$ as an RL agent and optimize it using Proximal Policy Optimization (PPO)~\cite{schulman2017proximal}.
When the reinforced partial convolution module $f_P$ takes the global feature map (i.e.,~the intermediate output of $f_G$) from more forward layers as the input, the computational cost to explore localities increases since there are more search locations with larger $H'$ and $W'$, and more subsequent convolution operations are needed to encode the selected locality to the same dimension of the global embedding $e^n_g$. On the other hand, in more forward layers, there may exist more useful information ignored in the global embedding. Therefore, as a trade-off between performance and computational cost, we assign $f_P$ after the $(n-1)$\mbox{-}th convolutional block $C^{n-1}_g$ of $f_G$. This way, the action space of $\pi$ is drastically reduced, and the regions become more information-intensive. It also becomes easier for $\pi$ to make decisions and achieve higher rewards. All the above-mentioned factors can help mitigate the sample efficiency problem of RL.

To better utilize global and local information, we design the state $s_{t}$ for $\pi$ 
to cover
two situations during selection:
\begin{equation}\label{state_definition}
s_{t}=\left\{\begin{matrix}
<\bar{e}_{g}^{n}, \varnothing> & t=1\\ 
<\bar{e}_{l,t-1}, h_{t-1}> & t>1
\end{matrix}\right.
\end{equation}
where $s_{t}$ denotes the state for the $t$\mbox{-}th step; $\varnothing$ denotes the empty set; $h_{t-1}$ denotes the hidden state from previous selection in the recurrent network. At first, we highlight the most helpful region for the global embedding. In the following steps, we keep previous selections as the hidden state and find the best locality for the current extracted representation.

Given the current state $s_{t}$, the policy network $\pi$ chooses a locating action $\mathit{loc}_{t}\sim \pi (\mathit{loc}_{t}|s_{t})$. $\mathit{loc}_{t}=\{i,j\}$ is a coordinate where $i\in [1,\frac{H'+2p-k}{q}+1]$ and $j\in [1,\frac{W'+2p-k}{q}+1]$. Then, we can obtain the region locality $e_{l,t}^{p}$ as: $e_{l,t}^{p}=\mathit{Conv}(\mathit{Crop}(\pi(s_{t}),e_{g}^{n-1}))$,
where $\mathit{Crop}$ is a Region of Interest (RoI) pool; $\mathit{Conv}$ is the convolution kernel for $f_{P}$. We use $\mathit{Crop}$ to align the output size during selection.

We repeat the procedure of selecting and extracting localities until the reward of $\pi$ exceeds a pre-defined threshold $\sigma$. Regions that have been visited will not be chosen again. The definition of the reward and the details of $\pi$ will be discussed in later section.
At this stage, $e_{l,t}^{p}$ is rough and insufficient for predicting attribute vectors, so we apply a locality extractor $C_{l}$ to further distill: $\bar{e}_{l,t}=C_{l}(e^{p}_{l,t})$.

To optimize the convolution kernels in $f_{P}$ and $C_{l}$, we apply a local predictor $f_{cl}$ to help train the kernels to effectively extract attribute-related localities by a locality loss $L_l$:
\begin{equation}
\begin{split}
    L_{l}&=\frac{1}{|\mathit{step}|}\sum_{t} L_{l,t}(\bar{a}_{l,t}, y)
    \\
    &=-\frac{1}{|\mathit{step}|}\sum_{t} \log\frac{\exp(f_{cl}(\bar{e}_{l,t})^{T}\phi(y))}{\sum_{\hat{y}\in Y^{S}}\exp(f_{cl}(\bar{e}_{l,t})^{T}\phi(\hat{y}))}
\end{split}
\end{equation}
where $|\mathit{step}|$ denotes the selection number. It may be insufficient to use the same ground-truth attributes to optimize the local subnet since we aim to capture diverse localities across steps. Therefore, we apply a maximum prediction loss $L_m$ to maximize locality diversity in next section.

\subsection{Joint Supervision Subnet}\label{joint subnet}
We conduct joint supervision over the global and local embeddings. Joint supervision consists of two losses: a joint prediction loss $L_{jnt}$ and a maximum prediction loss $L_{m}$. Both are evaluated by \textit{CrossEntropy}:
\begin{equation}\label{joint_prediction}
    L_{jnt}
    =-\log  \frac{\exp(f_{cj}(<\bar{e}_{g}^{n},\bar{e}_{l,1},...,\bar{e}_{l,t}>)^{T}\phi(y))}{\sum_{\hat{y}\in Y^{S}}\exp(f_{cj}(<\bar{e}_{g}^{n},\bar{e}_{l,1},...,\bar{e}_{l,t}>)^{T}\phi(\hat{y}))}
\end{equation}
\begin{equation}
    L_{m}
    =-\log \frac{\exp(\max_{i}(<\bar{a}_{g}^{n},\bar{a}_{l,1},...,\bar{a}_{l,t}>_{i})^{T}\phi(y))}{\sum_{\hat{y}\in Y^{S}} \exp(\max_{i}(<\bar{a}_{g}^{n},\bar{a}_{l,1},...,\bar{a}_{l,t}>_{i})^{T}\phi(\hat{y}))}
\end{equation}
where $f_{cj}$ is a fusion module to predict joint attributes $\bar{a}_{j}$ based on global and local visual embeddings of all steps; $\bar{a}_{m}$ is a vector composed of the maximum value in each dimension of the learned global and local attributes.

\textbf{Global-local cooperation}. We concatenate the global embedding with the corresponding localities to predict the attribute through $f_{cj}$. $f_{cj}$ is optimized by the loss $L_{jnt}$ to help the global and local embeddings collaborate better. Note that we use zero-padding to align the input of $f_{cj}$ since the lengths of action sequences differ across images.

\textbf{Locality diversity}. Our network aims to enable the local subnet to 
capture diverse localities.
Therefore, representations from different steps should 
emphasize
different parts of the attribute vectors. $L_{m}$ is designed to optimize the combinations of the most significant parts from global and local attribute embeddings. $L_m$ along with the locality loss $L_l$ jointly improve the locality diversity and discrimination. 

\subsection{Entropy-guided Policy Network}\label{policy_optimization}

The entropy-guided sampler $\pi$ is based on the global-local structure and joint feature learning.
We introduce information entropy as expert knowledge to help optimize the policy network.
A common obstacle of RL training is the sparse-reward problem, which occurs when the RL agent does not observe enough reward signals to reinforce its actions and then hinders the learning. 
Information entropy is a common tool to measure information quantity and can be used to guide the module towards informative regions that are more likely to contain useful localities, which, intuitively, can help alleviate the sparse-reward problem of RL.  

Given an arbitrary instance $(x,y,a)$, we obtain the corresponding locality sequence $\{\bar{e}_{g}^{n},\bar{e}_{l,1},...,\bar{e}_{l,t}\}$. 
During selection, we conduct the joint prediction for each step: $\bar{a}_{j,t} = f_{cj}<\bar{e}_{g}^{n},\bar{e}_{l,1},...,\bar{e}_{l,t}>$.
Then, we use the union prediction probability of the ground-truth label as the reward:
\begin{equation}
    r_{t} = \beta (\frac{exp(\bar{a}_{j,t}^{T}\phi(y))}{\sum_{\hat{y}\in Y^{S}}\exp(\bar{a}_{j,t}^{T}\phi(\hat{y}))} + \frac{exp(\bar{a}_{g}^{T}\phi(y))}{\sum_{\hat{y}\in Y^{S}}\exp(\bar{a}_{g}^{T}\phi(\hat{y}))})
\end{equation}
where $\beta$ is the entropy weight of instances. The weight $\beta$ is calculated as follows:
\begin{equation}
    \beta = \frac{\mathit{Entropy}(\mathit{Crop}(\mathit{loc}_{t}, e_{g}^{n-1}))}{\mathit{Entropy}(e_{g}^{n-1})}
\end{equation}
\begin{equation}
    \mathit{Entropy}(e)=-\sum_{i}\sum_{j}\sum_{k} p(e_{i,j,k})\log p(e_{i,j,k})
\end{equation}
where $i,j,k$ denote the coordinates; $\mathit{loc}_{t}$ denotes action for step $t$; $\mathit{Entropy}$ calculates the information entropy of the given region. We assess the entropy ratio of the selected region to the whole and use this ratio to represent the relative information richness. 
The entropy ratio can scale the prediction confidence to boost the policy network optimization.

\begin{table*}[ht]
\small
\centering
\setlength{\tabcolsep}{1.5pt}
\begin{tabular}{l|cccc|ccc|ccc|ccc|ccc}
\hline
\multirow{3}{*}{Method} & \multicolumn{4}{c}{ZSL} & \multicolumn{12}{c}{GZSL}\\\cline{2-17}
&SUN & CUB & aPY & AWA2 & \multicolumn{3}{c}{SUN} & \multicolumn{3}{c}{CUB} & \multicolumn{3}{c}{aPY} & \multicolumn{3}{c}{AWA2} \\\cline{2-17}
&T1 & T1 & T1 & T1 & U & S & H & U & S & H& U & S & H& U & S & H\\
\hline
\hline
\textbf{Non End-to-End}\\
SP-AEN\cite{chen2018zero} & 59.2&55.4&24.1&58.5 & 24.9&\textbf{38.6} & 30.3& 34.7 & 70.6 & 46.6 & 13.7 & 63.4 & 22.6 & 23.0 & 90.9 & 37.1\\
RelationNet\cite{sung2018learning}  & - &55.6&-&64.2 & - & - & - & 38.1 & 61.1 & 47.0 & - & - & -& 30.0 & \textbf{93.4} & 45.3 \\
PSR\cite{annadani2018preserving} & 61.4 & 56.0&38.4&63.8&20.8  &37.2  &26.7&24.6  &54.3   &33.9&13.5  &51.4  &21.4&20.7  &73.8  &32.3\\
PREN\cite{ye2019progressive} & 60.1 & 61.4 & - & 66.6 & 35.4 & 27.2 & 30.8 & 35.2 & 55.8 & 43.1 & - & - & - & 32.4 & 88.6 & 47.4 \\
\cline{1-17}
\textit{Generative Methods}\\
cycle-CLSWGAN\cite{felix2018multi} & 60.0 & 58.4 &- &67.3 &\textbf{47.9} & 32.4 & 38.7 & 43.8 & 60.6 & 50.8 & - & - & - & 56.0 & 62.8 & 59.2\\
f-CLSWGAN\cite{xian2018feature} & 58.6 & 57.7&-  & 68.2  & 42.6 & 36.6 & 39.4 & 43.7 & 57.7 & 49.7 & - & - & - & 57.9 & 61.4 & 59.6  \\
TVN\cite{zhang2019triple} & 59.3 & 54.9 & 40.9 & 68.8 & 22.2 & 38.3 & 28.1 & 26.5 & 62.3 & 37.2 & 16.1 & \textbf{66.9} & 25.9 & 27.0 & 67.9 & 38.6\\
SE-GAN \cite{pambala2020generative} & 61.8 & 60.8 & - & 68.8 & 44.7 &37.0 &\textbf{40.5} &48.4 &57.6 &52.6 &- &- &- &55.1 &61.9&58.3\\
Zero-VAE-GAN\cite{gao2020zero} & 58.5 & 51.1 & 34.9 & 66.2 & 44.4&30.9&36.5&41.1&48.5&44.4&30.8&37.5&33.8&56.2&71.7&63.0\\
\hline
\hline
\textbf{End-to-End}\\ 
QFSL\cite{song2018transductive}&56.2&58.8&-&63.5&30.9  &18.5  &23.1 &33.3  &48.1 &39.4&-&-&-&52.1 & 72.8  &60.7\\
SGMA\cite{zhu2019semantic}&- &71.0&-&68.8& - & - & - & 36.7 & 71.3 & 48.5 & - & - & - & 37.6 & 87.1 & 52.5\\
LFGAA\cite{Liu_2019_ICCV}&61.5&67.6&-&68.1& 20.8 & 34.9 & 26.1 & 43.4 & \textbf{79.6} & 56.2 & - & - & - & 50.0 & 90.3 & 64.4\\
AREN\cite{xie2019attentive}&60.6&71.5&39.2&67.9& 40.3 & 32.3 & 35.9 & 63.2 & 69.0 & 66.0 & 30.0 & 47.9 & 36.9 & 54.7 & 79.1 & 64.7\\
SELAR-GMP\cite{yang2020simple} & 58.3 & 65.0 & - & 57.0 & 22.8 & 31.6 & 26.5 & 43.5 & 71.2 & 54.0 & - & - & - & 31.6 & 80.3 & 45.3\\
APN\cite{xu2020attribute}&60.9&71.5&-&68.4& 41.9 & 34.0 & 37.6 & 65.3 & 69.3 & 67.2 & - & - & - & 56.5 & 78.0 & 65.5\\
\hline
\textbf{Ours} 
ERPCNet  & \textbf{63.3} & \textbf{72.5}&  \textbf{43.5} &\textbf{71.8} &47.2 & 31.9 & 38.1 & \textbf{67.1} & 69.6 & \textbf{68.4} & \textbf{32.7} & 49.3 & \textbf{39.3} & \textbf{59.1} & 82.0  & \textbf{68.7}\\
\hline
\end{tabular} 
\caption{Main experiments.}
\label{table main}
\end{table*}

Finally, we can optimize the following loss function according to the work of Schulman et al.~\cite{schulman2017proximal}: $\max_{\pi}\mathbb{E}[\sum_{t}\gamma^{t-1}r_{t}]$,
where $\gamma$ denotes a discount parameter. The detailed 
optimization  
is
given in \textit{Appendix~\ref{PPO}}.

\subsection{Training and Inference}
We train our model in an end-to-end manner. To prevent overfitting, we set a maximum step number $T$ and 
halt
the selection 
once the reward exceeds the threshold $r_t \geq \sigma$ ($1\leqslant t \leqslant T$) or after $T$ steps. 

\textbf{Training} We use a two-stage strategy to maximize the prediction capability with the fewest locality proposals. At \textbf{stage \uppercase\expandafter{\romannumeral1}}, we train the model to predict correctly for an arbitrary sequence of local regions.
Instead of using $\pi$, we randomly select local regions at each step without early-stopping. 
Then, we optimize the rest of the model by minimizing the overall loss: $L_{\emph{erpc}} = L_g + L_l + L_j + L_m$.
At \textbf{stage \uppercase\expandafter{\romannumeral2}}, we fix the modules' parameters trained in Stage \uppercase\expandafter{\romannumeral1} and use $\pi$ to select locations (with early-stopping). Then, we apply PPO to optimize $\pi$ to pick the most discriminative localities.

\textbf{Inference}
We use the union of the global and local prediction for inference: $\bar{a} = \bar{a}_{j,t}+\bar{a}_{g}$.
For ZSL, given an image $x$, the model extracts global information and then performs locality search iteratively until 
the termination condition. 
During inference, the model considers the predicted label as ground-truth to calculate the reward. Then, we take the class with the highest compatibility as the final prediction: $ y^{U}=\argmax \bar{a}_{\hat{y} \in Y^{U}}^{T}a_{\hat{y}\in Y^{U}}$.
For GZSL, since both seen and unseen classes may occur during testing, there exists a strong bias toward seen classes. To alleviate the bias, we adopt Calibrated Stacking (CS)~\cite{chao2016empirical} to decrease the confidence of seen classes by a constant. The final prediction is: $ y^{U\cup S}=\argmax <\bar{a}_{\hat{y}\in Y^{U}}^{T}a_{\hat{y}\in Y^{U}},\bar{a}_{\hat{y}\in Y^{S}}^{T}a_{\hat{y}\in Y^{S}}-\delta>$,
where $\delta$ is a pre-defined parameter.

\section{Experiments}\label{exp}

We conduct 
experiments on four benchmark datasets for both ZSL and GZSL: SUN~\cite{patterson2012sun}, CUB~\cite{welinder2010caltech}, aPY~\cite{farhadi2009describing}, and AwA2~\cite{xian2019zero}.
SUN and CUB are fine-grained datasets, containing 14,340 images from 717 scene classes with 102 attributes and 11,788 images from 200 bird species with 312 attributes, respectively; aPY contains 15,339 images from 32 classes with 64 attributes, where images are from two distinct main types (buildings and animals); AwA2 is a large coarse-grained dataset comprising 37,322 images from 50 diverse animals with only 85 attributes. We adopt Proposed Split (PS)~\cite{xian2019zero}, which is commonly used to avoid unseen data leak, to divide datasets into seen/unseen classes. 

We adopt Resnet101~\cite{he2016deep} pretrained on ImageNet~\cite{deng2009imagenet} as the backbone (i.e.,~the global extractor $f_G$) and divide $f_G$ into blocks $\{C_g^1,C_g^2,\dots,C^n_g\}$ following~\cite{he2016deep}.
$C_l$ shares the same structure and initial parameters with $C^n_g$ but with different parameters after optimization.
At Stage \uppercase\expandafter{\romannumeral1}, we use SGD~\cite{bottou2010large} with image size of $224\times224$, momentum of 0.9, weight decay of $10^{-5}$, and a learning rate of $10^{-3}$.
The learning rate decays by 0.1 every 30 epochs.
At Stage \uppercase\expandafter{\romannumeral2}, we use Adam~\cite{kingma2014adam} to optimize $\pi$ with a learning rate of $3\times10^{-4}$ and $\gamma$ of 0.99. The maximum step $T$ is set to be 10 for AwA2, and 6 for other datasets. 
More parameters and network architecture 
are given
in 
\textit{Appendix \ref{appendix implement}}.

\subsection{Comparisons with Baselines and Ablation Study}\label{exp main}
\textbf{ZSL}: We compare our method with two groups of state-of-the-art methods: non-end-to-end methods (including embedding methods and generative methods) and end-to-end methods. We evaluate the methods by average per-class Top-1 (T1) accuracy to mitigate the influence of class imbalance.
Results are shown in Table~\ref{table main}. For competitors, we use the accuracy reported in the original papers. Since APN~\cite{xu2020attribute} additionally uses group side information (besides class labels), we list the results of its without-group version to make a fair comparison.

Table~\ref{table main} shows that our method consistently outperforms other models (and especially other end-to-end methods) by a large margin. In particular, ERPCNet outperforms the second-best method by 1.5\%, 1\%, 2.6\%, and 3\% on SUN, CUB, aPY and AWA2, respectively. The performance gain on SUN and CUB (which contains fewer images for each class) are not as significant as on AwA2 and aPY. 

\textbf{GZSL}: Following~\cite{xian2019zero}, we evaluate the average per-class accuracy on seen classes (denoted by $S$), unseen classes (denoted by $U$), and their harmonic mean (defined as $H=\frac{2US}{U+S}$) in the GZSL setting.
Table~\ref{table main} shows that our model outperforms all other embedding approaches, especially on the aPY and AwA2 datasets, yielding 2.4\% and 3.2\% improvements of $H$, respectively. The results demonstrate that our model can transfer knowledge from seen classes to unseen classes successfully. 

GZSL needs to classify both seen and unseen classes, thus there exist a strong bias towards seen classes during testing. Generative methods can, to some extent, address the problem naturally by synthesizing instances for unseen classes. This explains why generative methods perform better than non-generative methods in GZSL. Interestingly, our model's performance is comparable to or better than generative models, which demonstrates our model's generalization ability.

\begin{table}
	\centering
	\small
\begin{tabular}{lcccc}
\hline
Method & SUN & CUB & aPY & AwA2  \\ 
\hline
\hline
GlobalNet &  61.3 & 68.1& 39.4 &66.9\\
PCNet   & 62.8 & 71.2&  41.8&69.6\\
RPCNet & 63.3 &  72.0& 43.5&71.6\\
ERPCNet & 63.1 & 72.5&  43.5&71.8\\
\hline
\end{tabular} 
\caption{Ablation in ZSL.}
\label{table_zsl_ablation}
\end{table}

\textbf{Ablation study}: We also compare with GlobalNet (classification using only global subnet), PCNet (randomly selecting locality), and RPCNet (ERPCNet without entropy guidance) as ablations in ZSL. 
Our proposed ERPCNet is effective on four benchmark datasets, demonstrated by improvement of T1 by up to 2.0\%, 4.4\%, 4.1\% and 4.9\% on SUN, CUB, aPY, and AwA2, respectively, when compared with GlobalNet (shown in Table~\ref{table_zsl_ablation}).
The improvement derives from three aspects: 1) the proposed partial convolution to extract and incorporate local information (proved by the superiority of PCNet over GlobalNet), 2) the use of RL to progressively select localities (confirmed by the advantage of RPCNet over PCNet), and 3) the guidance of entropy (demonstrated by comparing ERPCNet with RPCNet).

\subsection{Efficacy of entropy-guided reinforcement learning}\label{sec:efficacy}

\begin{figure}[h]
\centering 
\begin{subfigure}{0.2\textwidth}
\centering
\includegraphics[width=\textwidth]{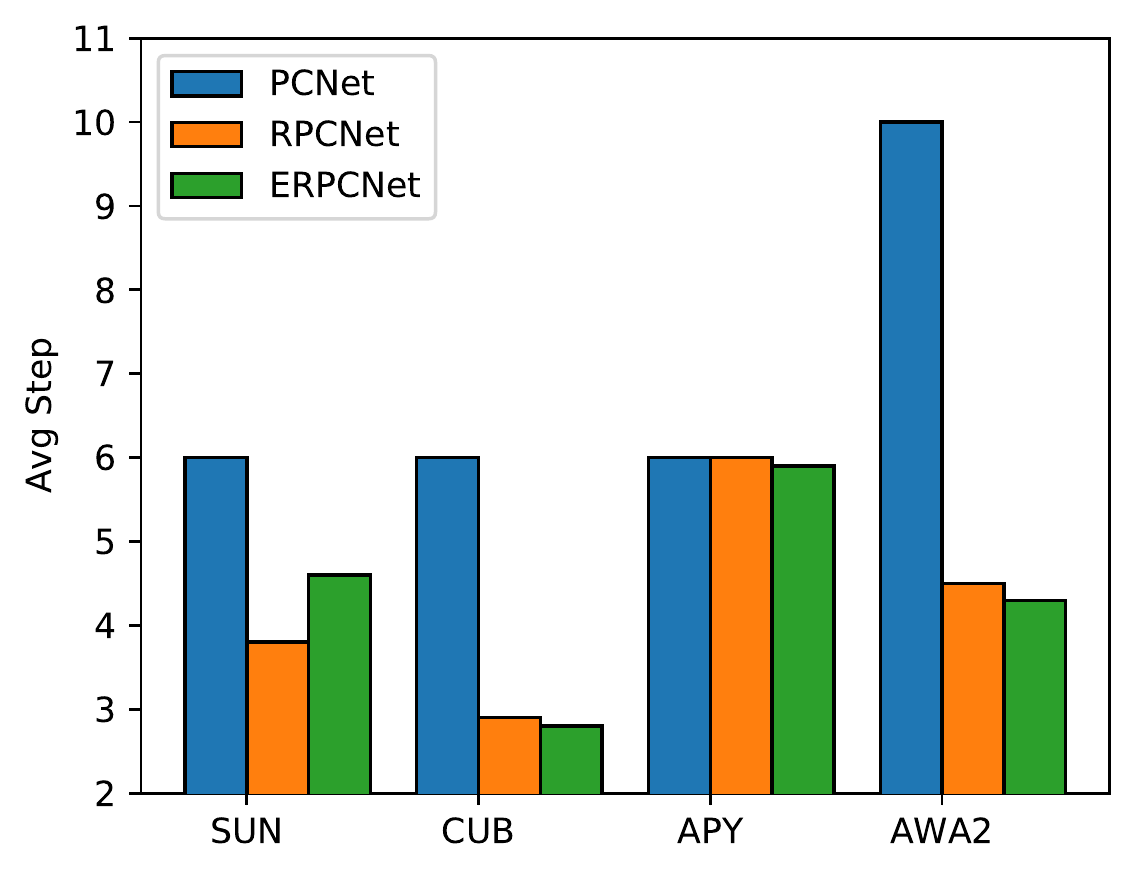}
\caption{Average Step.}
\end{subfigure}\hfil 
\begin{subfigure}{0.22\textwidth}
\centering
\includegraphics[width=\textwidth]{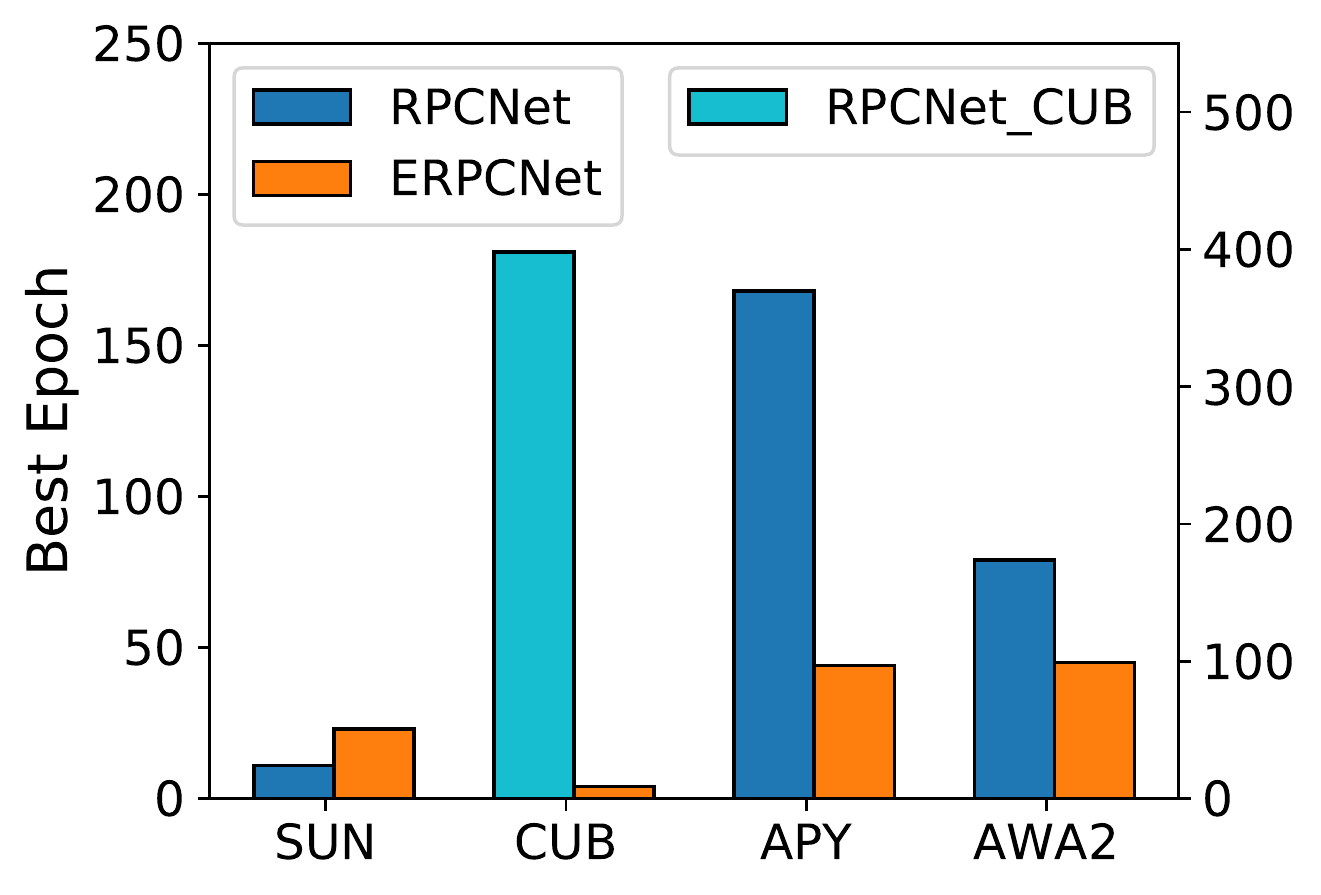}
\caption{Best epoch.}
\end{subfigure}\hfil 
\begin{subfigure}{0.2\textwidth}
\centering
\includegraphics[width=\textwidth]{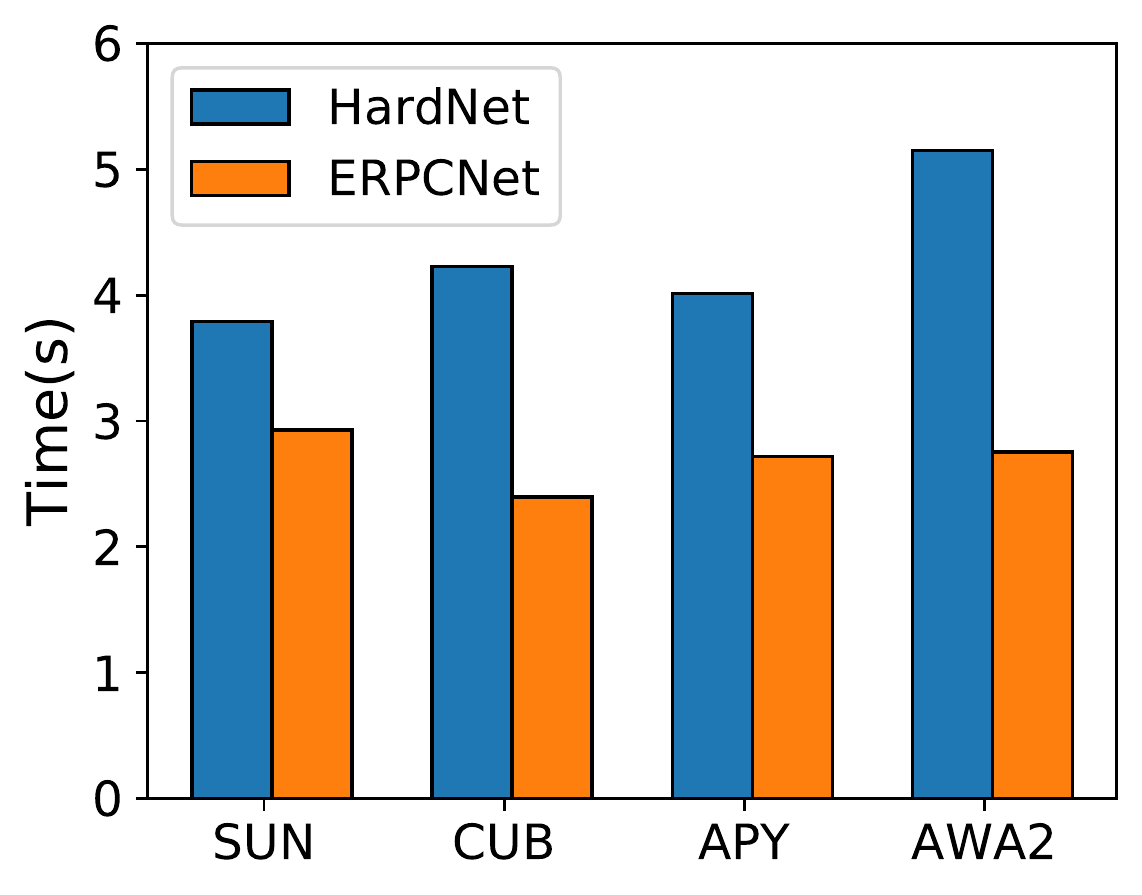}
\caption{Average train time.}
\end{subfigure}\hfil 
\begin{subfigure}{0.2\textwidth}
\centering
\includegraphics[width=\textwidth]{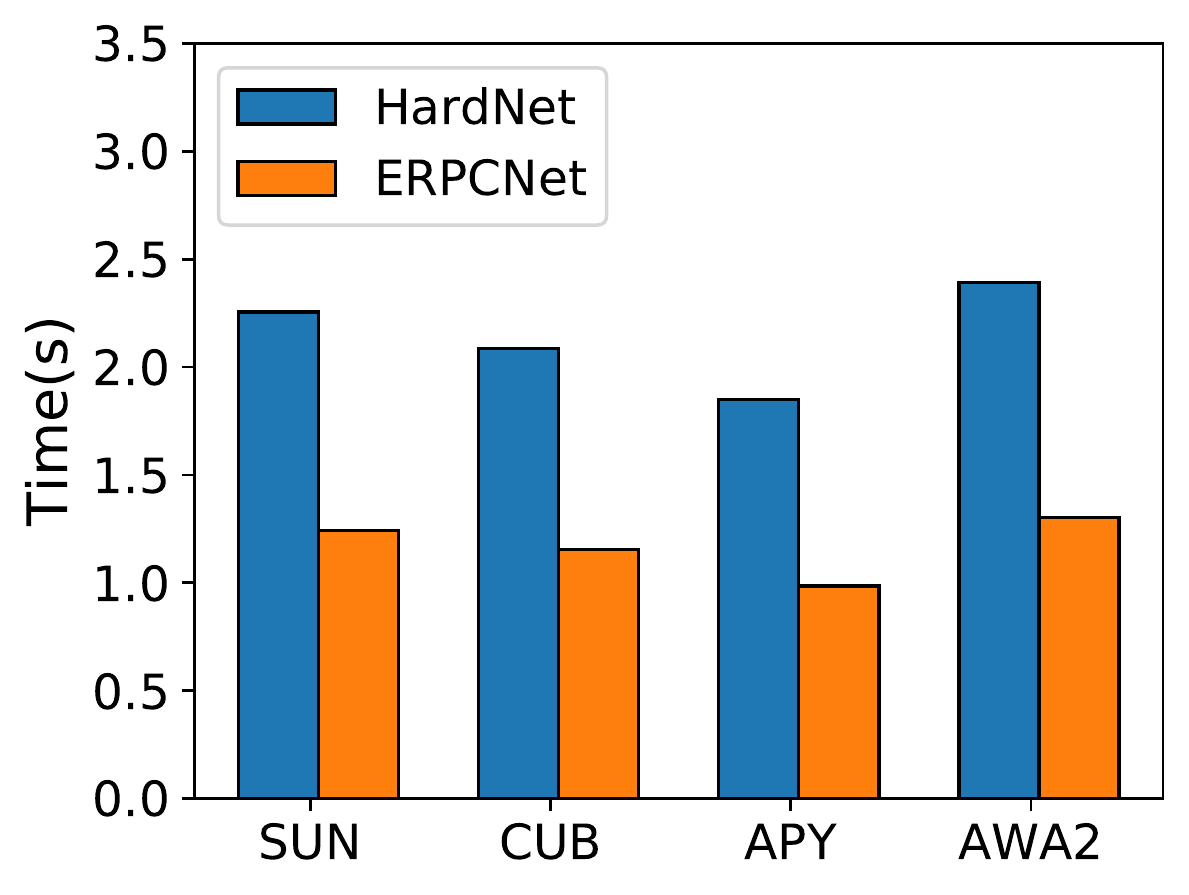}
\caption{Average test time.}
\end{subfigure}\hfil 
    \caption{(a)-(b) Comparison between using and not using entropy-guidance. (c)-(d) Efficiency analysis (unit: s).}
\label{efficacy and efficiency}
\end{figure}

Figures~\ref{efficacy and efficiency}(a)-(b) shows the average terminating steps and best epochs (i.e.,~where the model achieves the best accuracy) of RPCNet and ERPCNet on the four datasets. Both RPCNet and ERPCNet take fewer steps than PCNet (6/10 steps) but achieve higher accuracy, indicating the effectiveness of the reinforced module $\pi$.
Entropy-guided RL can largely 
decrease the number of epochs required to obtain the best performance on CUB, aPY and AwA2. Besides, we draw the Acc-epoch curves in \textit{Appendix} to further prove that entropy guidance can boost RL training.
Also, entropy knowledge can slightly reduce the steps during testing.
Entropy knowledge does not work well on the SUN dataset. We analyze the value ranges of the entropy weight $\beta$ and find that $\beta$ on SUN (on average, 1.09) is slightly smaller than on other datasets (on average, 1.12), which may impair the results.

\subsection{Efficiency of partial convolution}

To examine the efficiency of partial convolution, we compare our model against using hard attention~\cite{xu2015show} to explore localities (denoted by HardNet). Hard attention finds important image patches and extracts localities from the cropped images. We train two feature extractors sharing the same structure with our $f_G$ to learn from the original images and the cropped patches, respectively. We also adopt a PPO agent $\pi'$ for HardNet optimization. 
Since the size ($H'\times W'$) of the feature map for partial convolution is $14\times14$, with the kernel size being $5\times 5$ and the HardNet input image size being $224 \times 224$, we set the patch size in HardNet to $80\times80$ proportionally.
The average training and testing time of a single instance for the optimization of $\pi$ and $\pi'$ is shown in Figures~\ref{efficacy and efficiency}(c)-(d), and our model consumes around $2/3$ and $1/2$ of the HardNet training/testing time, respectively. The results demonstrate the 
efficiency of our partial convolution design. Integrating RL with convolution reduces the action space from any location in $224\times224$ images to $4\times4$, thus reducing the time cost.

\subsection{Hyper-parameters}\label{exp hyper}

\begin{figure}{r}
\centering 
\begin{subfigure}{0.21\textwidth}
\centering
\includegraphics[width=\textwidth]{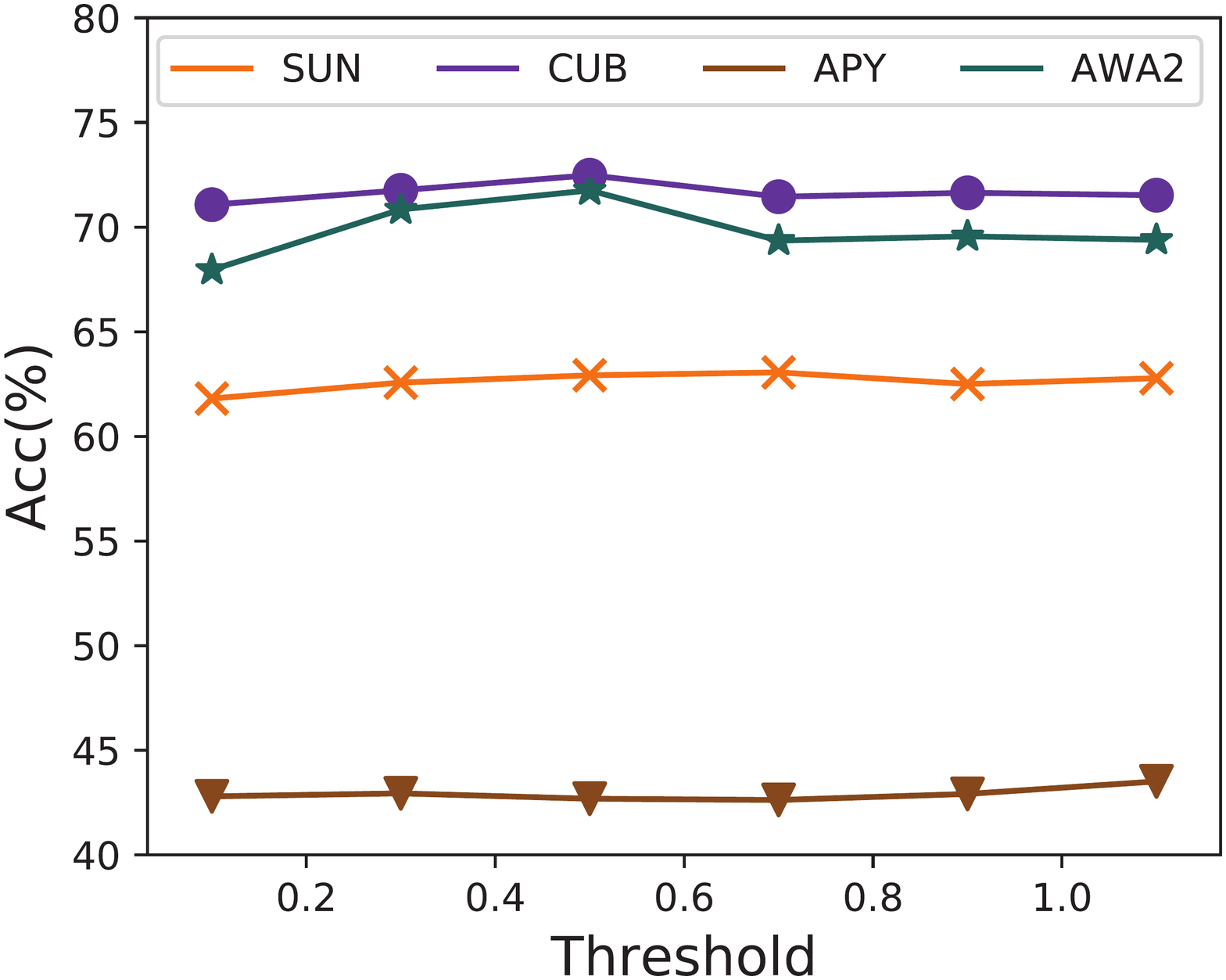}
\caption{Threshold analysis.}
\end{subfigure}\hfil 
\begin{subfigure}{0.22\textwidth}
\centering
\includegraphics[width=\textwidth]{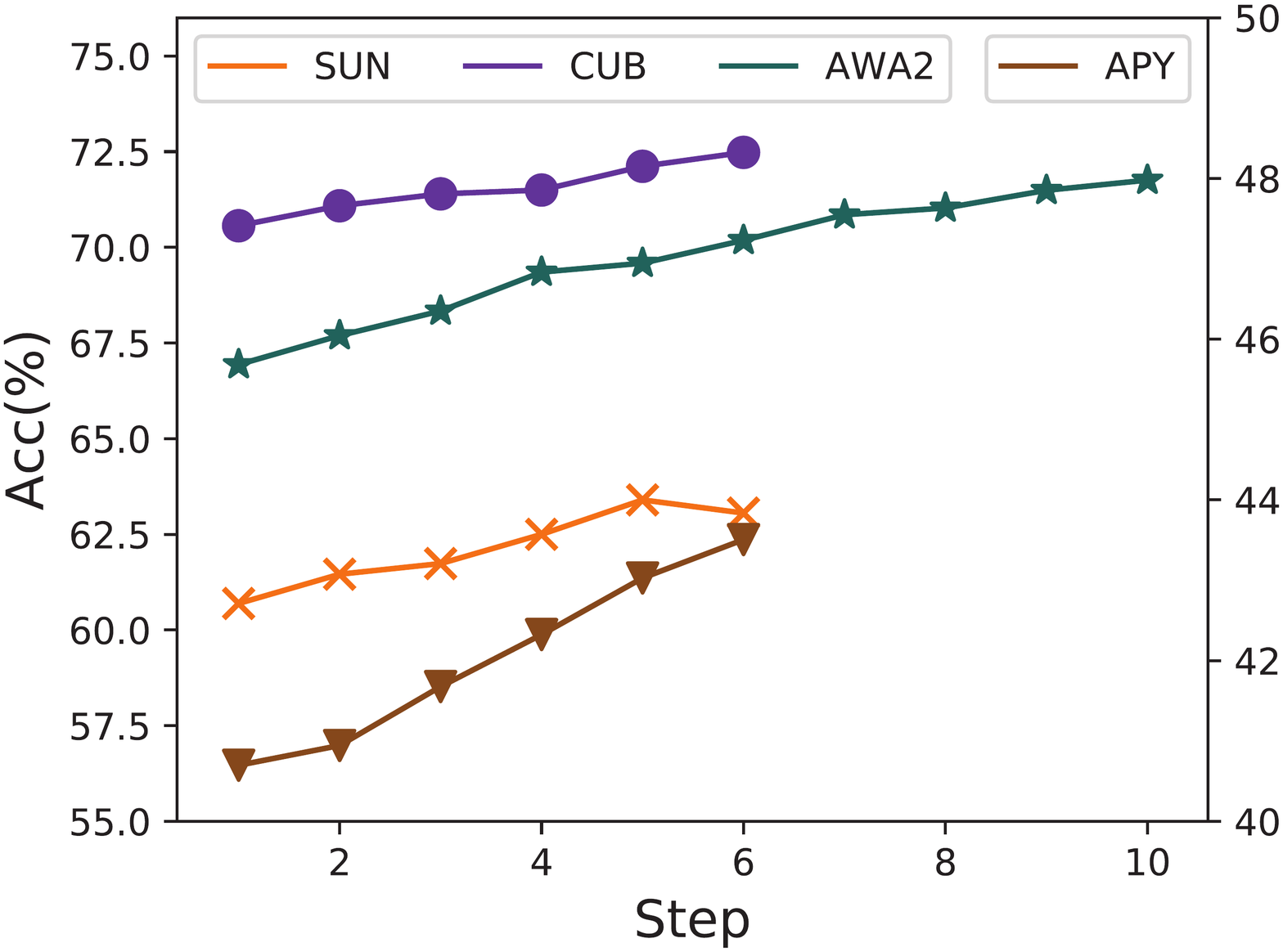}
\caption{Step-acc curve.}
\end{subfigure}\hfil 
\caption{Hyper-parameter analysis.}
\label{hyper-parameter}
\end{figure}

\begin{figure}[!htb]
\centering 
\includegraphics[width=\linewidth]{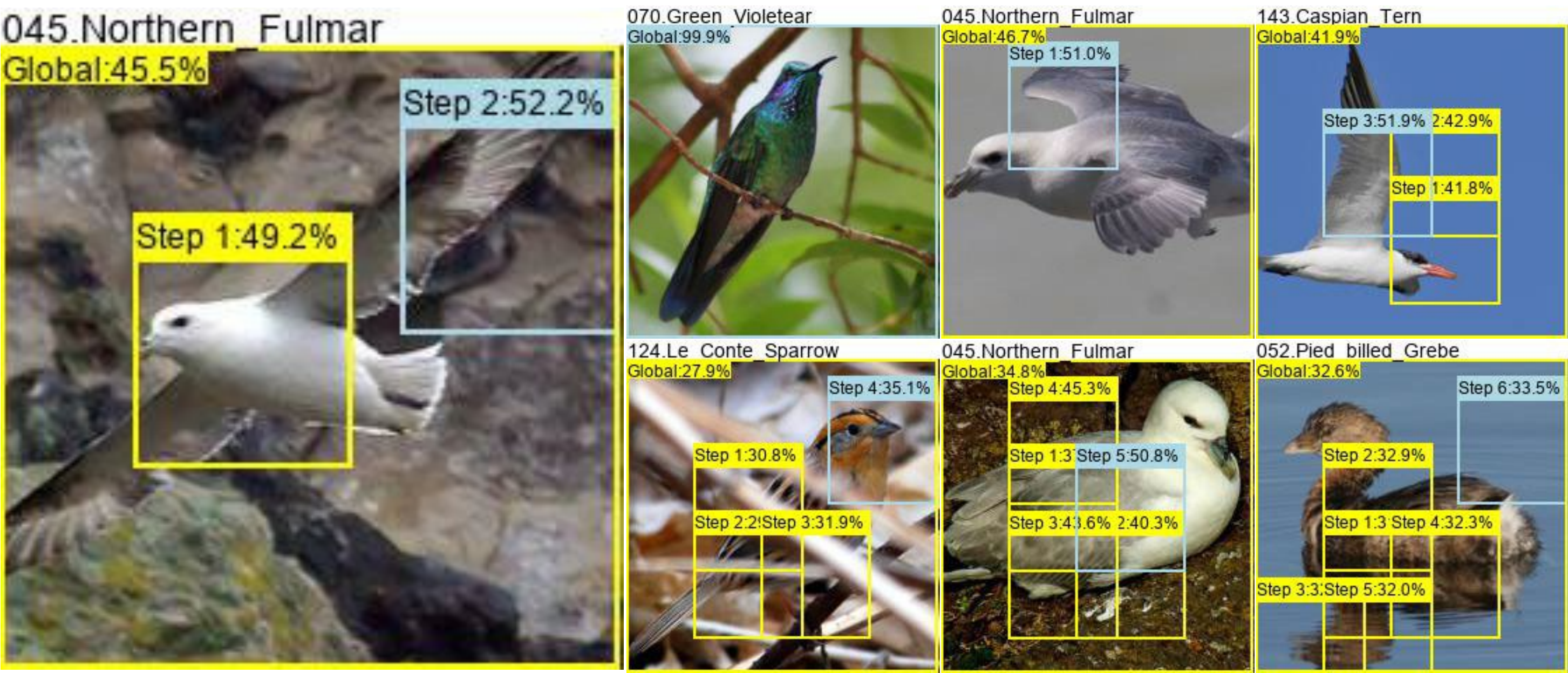}
\caption{Progressive locality selection on CUB. The labels above the boxes denote the step index and the prediction confidence after this selection. The box color indicates the prediction correctness (blue: correct; yellow: wrong).}
\label{cub_locality_visualization}
\end{figure} 

\textbf{Threshold $\sigma$ of $\pi$}: We show the performance of ZSL varying $\sigma$ from 0.1 to 1.1 with a step of 0.2 in Figure~\ref{hyper-parameter}(a).
The results are stable when $\sigma$ is over 0.7 and slightly influenced by $\sigma$ when $\sigma\in[0.1,0.5]$.

\textbf{Step-acc curve}: We fix the maximum steps $T$ to be 6 on three datasets (SUN, CUB, and aPY) and 10 on AwA2. The step-accuracy curves in Figure~\ref{hyper-parameter}(b) show the accuracy increases as more steps are performed, and the improvement tends to be subtle after five steps or even diminishes on SUN. The results indicate that the locality incorporation benefits classification, but introducing excessive locality could be harmful. Analysis for RPCNet is provided in \textit{Appendix}~\ref{app_exp}.

\subsection{Progressive Process Visualization}

Figures~\ref{cub_locality_visualization} visualize instances that are easily predictable with the global representation and ones that can only be correctly classified with progressive localities on CUB. For easier understanding, we project the selected locality in the abstract hierarchies $loc_t$ ($1\leqslant t \leqslant T$) into the original image-level and use bounding boxes to represent the locations.
We find that green violetears 
can be easily classified with 
probability of 99.9\%, due to their 
distinctiveness
from other species.
When classifying similar bird species, the locality detector can gradually increase the probability of correct labels by locating the regions of wing, neck, head, etc. to highlight the birds' discriminative characteristics.
This indicates that our model can progressively pick up the best locality to help distinguish similar or diverse objects effectively.
We also visualize selection procedure on SUN in \textit{Appendix}~\ref{app_selection}.

Besides, we investigate the failure modes of the RL module to find out when RL will fail to find helpful localities in ~\textit{Appendix}~\ref{app_fail}.
We also visualize the distributions of the global and the union embedding on AwA2 by t-SNE~\cite{van2008visualizing} to demonstrate our model's ability to learn discriminative embeddings in \textit{Appendix}~\ref{app_embedding}.

\section{Related work}

\textbf{Zero-Shot Learning (ZSL).} ZSL aims to classify classes 
not seen
during training. A typical strategy is to view ZSL as a visual-semantic embedding problem, which reduces to designing an appropriate projection that maps visual~\cite{ye2019progressive,sung2018learning,chen2018zero} and/or semantic features~\cite{zhang2017learning,shigeto2015ridge} to a latent space, where ZSL measures the compatibility score of the latent representation for classification. For example, Ye et al.~\cite{ye2019progressive} design an ensemble network to learn an embedding  from the same extracted features to diverse labels. Several recent efforts~\cite{zhang2019triple,gao2020zero,felix2018multi,li2020grounding} convert ZSL to traditional supervised classification by exploring generative models to generate samples for unseen classes.

More related to our work, end-to-end models are proposed for better image representation~\cite{song2018transductive,zhu2019semantic,Liu_2019_ICCV,bustreo2019enhancing,xie2019attentive,xu2020attribute}. LFGAA~\cite{Liu_2019_ICCV} uses instance-based attribute attention to disambiguate semantic characteristics. Xie et al.~\cite{xie2019attentive} combine two branches of the multi-attention module to facilitate embedding learning and attribute prediction. 
However, multi-attention 
discovers a fixed number of localities independently while neglecting their region relations, thus restricting the attention weights to the global level. In contrast, ERPCNet can uncover refined local regions progressively while preserving attribute relevance and inherent correlations.

\textbf{Locality and representation learning.} 
Locality has been extensively investigated for better representation~\cite{zhu2019semantic,xu2020attribute,sylvain2019locality,hjelm2018learning}. Annotation-based methods~\cite{akata2016multi,elhoseiny2017link,ji2018stacked} leverage extra annotations in the form of ground-truth bounding boxes to extract local information or train local detectors. Weakly-supervised methods~\cite{huynh2020compositional,sylvain2019locality,hjelm2018learning,liu2020spectrum} can avoid labor-intensive annotations. \cite{xie2019attentive,wang2015multiple,zhang2016picking,huynh2020fine} adopt multi-attention to independently search important regions and treat them equally. Xu et al.~\cite{xu2020attribute} propose a prototype network to improve localities by concentrating on semantic groups. Wang et al.~\cite{wang2020glance} use a patch proposal network to focus on discriminative regions and remove spatial redundancy.

{\bf Summary.} Our model differs from previous studies on three aspects. 1) We first propose a new reinforced framework to find localities in ZSL and jointly learn zero-shot recognition, reinforced locality exploration, and global-local representations in an end-to-end manner. 2) We design entropy as guidance to identify information-rich regions in order to accelerate the training phase and alleviate sparse-reward problems. 3) We propose reinforced partial convolution to discover localities, which converges faster and reduces the computational cost.

\section{Conclusion}
We propose an Entropy-guided Reinforced Partial Convolutional Network (ERPCNet) to gain better global-local representations in Zero-Shot Learning (ZSL).
We perform partial convolution by incorporating a reinforced region sampler with a convolution kernel to dynamically find and learn localities as complements for the global representation.
We further introduce entropy knowledge into the reward design to guide the model toward informative regions.
We evaluate our model through extensive experiments against state-of-the-art methods in both ZSL and GZSL settings on four benchmark datasets, where the results demonstrate the superior performance and robustness of ERPCNet for global-local representation learning. Our comprehensive ablation studies show our model's effectiveness in locality exploration and efficiency in the training/testing of the reinforced module.
In the future, 
we will further explore augmenting other convolutional networks with ERPCNet in a plug-and-play manner to boost their performance.

\bibliography{aaai22}

\newpage
\appendix

\section{Proximal Policy Optimization}\label{PPO}
Our reinforcement module is implemented by an Actor-critic network, which consists of an actor $\pi$ and a critic $V$. The critic $V$ aims to estimate the state value~\cite{schulman2017proximal}. The detailed module architecture is shown in Section Architecture Implementation.

During the training process of the reinforcement module, we sample actions following $loc\sim \pi(loc|s_{t})$ to optimize the policy network, where $s_{t}$ denotes the state for the $t$-th step by maximizing the following rewards:
\begin{equation}
    \max_{\pi}\mathbb{E}[\sum_{t}\gamma^{t-1}r_{t}]
\end{equation}
where $\gamma=0.99$ is a pre-defined discounted parameter and $r_{t}$ denotes the reward. According to the work of Schulman et al.~\cite{schulman2017proximal}, the optimization problem can be addressed by a surrogate objective function using stochastic gradient ascent:
\begin{equation}
    L_{t}^{CPI}=\frac{\pi(loc|s_{t})}{\pi_{old}(loc|s_{t})}\hat{D}_{t}
\end{equation}
where $\pi_{old}$ and $\pi$ represent the before and after updated policy network, respectively. $\hat{D}_{t}$ is the advantages estimated by an Actor-critic network $V$ by:
\begin{equation}
    \hat{D}_{t}=-V(s_{t})+\sum_{t\leq i \leq T} \gamma^{i-t}r_{t}
\end{equation}
where $T$ denotes the maximum length of the action sequence. The policy network usually gets trapped in local optimality via some extremely great update steps when directly optimizing $L^{CPI}$, so we optimize a clipped surrogate objective:
\begin{equation}
    L_{t}^{CLIP}=min\{\frac{\pi(loc|s_{t})}{\pi_{old}(loc|s_{t})}\hat{D}_{t},Clip(\frac{\pi(loc|s_{t})}{\pi_{old}(loc|s_{t})})\hat{D}_{t}\}
\end{equation}
where $Clip$ is the operation that clips input to $[1-\epsilon, 1+\epsilon]$. We set $\epsilon=0.2$ in our experiments.

Then, to further promote the exploration of policy and the performance of $V$, we take the following loss function as the final optimization goal:
\begin{equation}
    \max_{\pi,V}\mathbb{E}_{x,t}[L_{t}^{CLIP}-\alpha_{1}MSE(V(s_{t},\sum_{t\leq i \leq T} \gamma^{i-t}r_{t})+\alpha_{2}S_{\pi}(s_{t})]
\end{equation}
where $\alpha_{1}=0.5$, $\alpha_{2}=0.01$, $MSE$ is the mean square error loss, and $S_{\pi}(s_{t})$ denotes the entropy bonus~\cite{williams1992simple,schulman2017proximal}.

\section{More Implementation Details}\label{appendix implement}
All algorithms are implemented in Pytorch 1.7.0 and complied by GCC 7.3.0. The system is Linux 3.10.0 and the GPU type is GP102 TITANX. The cuda version is 10.0.130. The stop threshold $sigma$ of $\pi$ is set to be 0.7, 0.5, 1.1 and 0.5 for SUN, CUB, aPY, and AwA2, respectively.
For GZSL, the factor $\delta$ of CS is set to 0.2, 0.8, 0.5, and 0.5 for SUN, CUB, aPY, and AwA2, respectively.

\subsection{Architecture Implementation}\label{Architecture Implement}
Our model relies on the convolution layer and fully connected layer. $FC(n)$ represent a fully-connected layer with output size $n$. We use the same network structure for all four benchmark datasets yet different parameters for dropout layers. In the following, we introduce the detailed network architecture of global subnet, local subnet, and other prediction layers, respectively.

First, we introduce the common setting for the layers. We use adaptive average pool (AdaptiveAvgPool) with output size $1\times 1$, rectified linear activation function (ReLU) with default parameter and sigmoid activation function with default parameter for each module. In the global subnet, which is composed of $f_{G}$ followed by an adaptive average pool, the input is the cropped image with the size of $3\times 224\times 224$. We use the pre-trained ResNet-101~\cite{he2016deep} for $f_{G}$ and set the output size to $1\times 1$ for AdaptiveAvgPool, where the output size of the global subnet is $N\times 2048$, and $N$ denotes batch size.

The local subnet consists of a partial convolution module $f_{P}$ and a convolution layer module $C_{l}$. To keep the same structure as the global subnet, we use the last block of ResNet-101 as $C_{l}$. As for the reinforced partial convolution module, $f_{P}$ contains a policy network $\pi$ and a convolution kernel (size $5\times 5$ and stride step 3). $\pi$ shares the same state encoder structure $f_{E}$ with the state value estimator $V$. The state structure $f_{E}$ is a recurrent network as follows:
\begin{equation}
    f_{E}=<FC(1024)-ReLU-FC(256)-ReLU-GRU)>
\end{equation}
where $GRU$ denotes a gated recurrent unit with input size 256 and hidden size 256. Then, we design $\pi$ and $V$ by:
\begin{equation}
\begin{gathered}
    \pi=<f_{E}-FC(|Action|)-Sigmoid>\\
    V=<f_{E}-FC(1)>
\end{gathered}
\end{equation}
where $|Action|$ denotes the action dim.

In respect of the prediction layers, $f_{cj}=<FC(2048)-Dropout-FC(|A|)-Dropout>$, $f_{cg}$ and $f_{cl}$ share the same structure as $<FC(|A|)-Dropout>$, where $|A|$ denotes the attribute vector dim and $Dropout$ is the dropout layer. The dropout layer parameters for CUB, aPY, AwA2 and SUN are 0.5, 0.5, 0 and 0 respectively.

\section{More Experiments}
\label{app_exp}

\subsection{Training Convergence Analysis}

\begin{figure}[htp]
\centering 
\begin{subfigure}{0.22\textwidth}
\centering
\includegraphics[width=\textwidth]{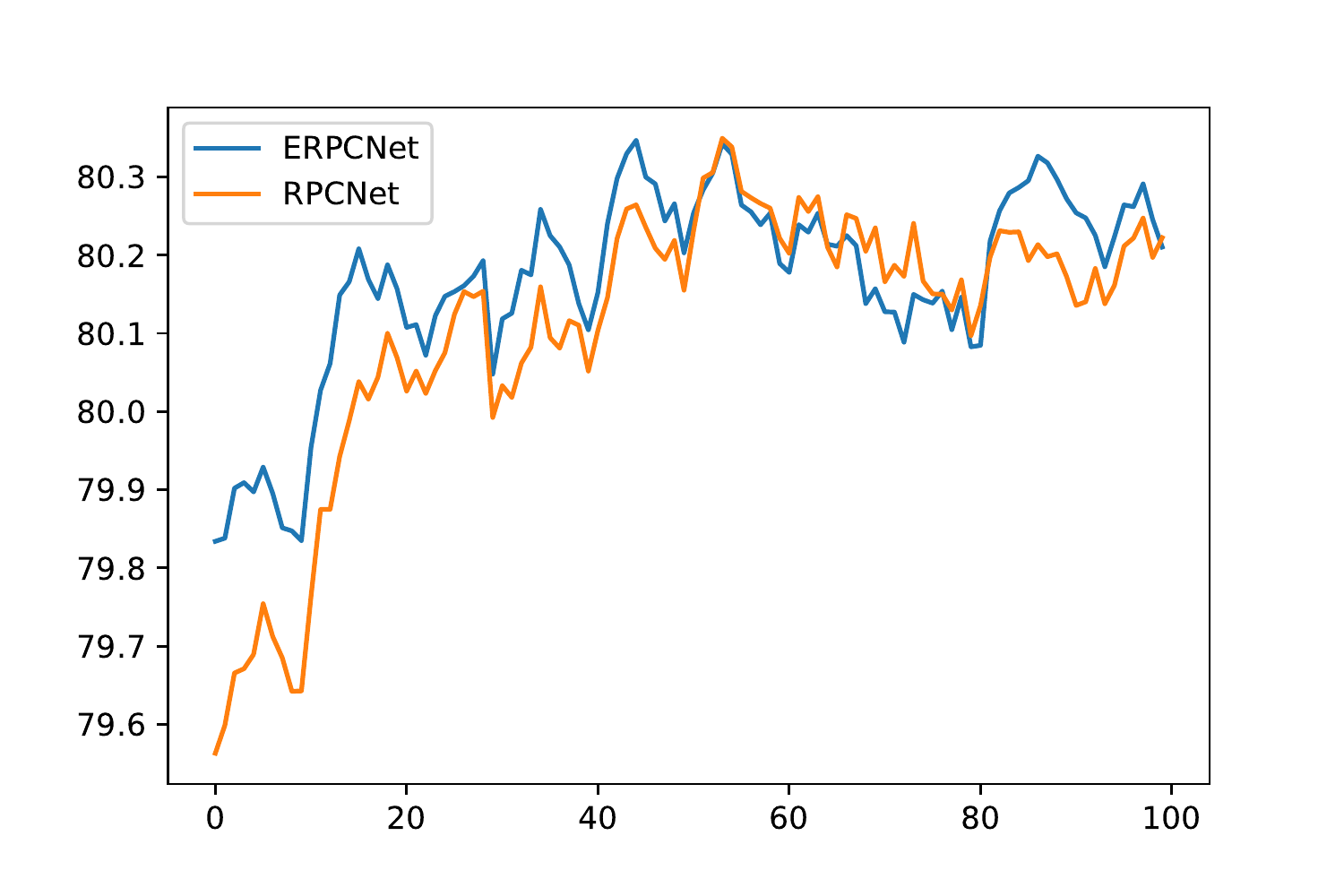}
\caption{CUB.}
\end{subfigure}\hfil 
\begin{subfigure}{0.22\textwidth}
\centering
\includegraphics[width=\textwidth]{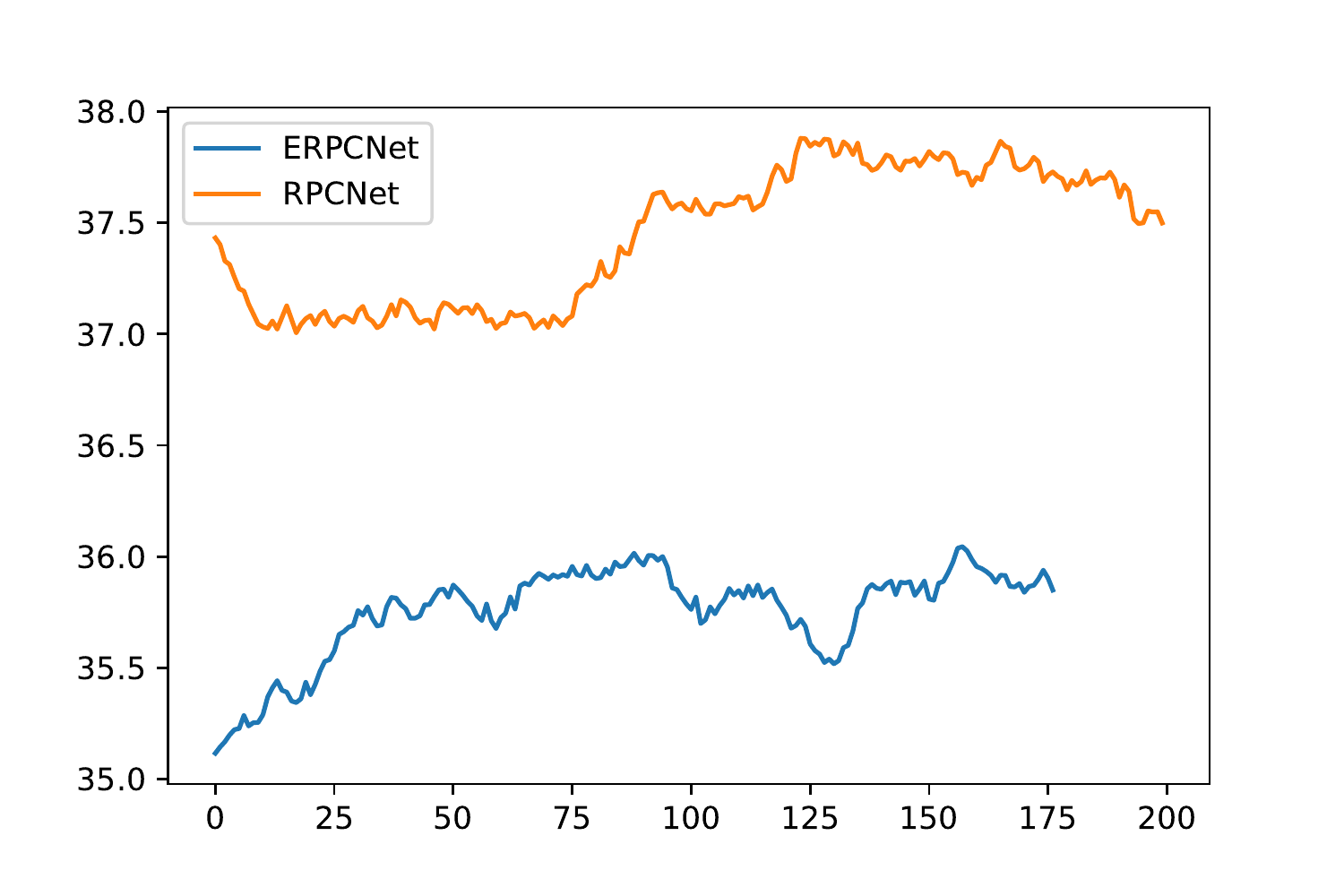}
\caption{SUN.}
\end{subfigure}\hfil 
    \caption{Acc-epoch curves.}
\label{appendix_converge}
\end{figure}

To further demonstrate our claim that the proposed entropy guidance can accelerate training convergence and improve performance, we show how the training accuracy changes as more epochs are performed on CUB and SUN in Figure~\ref{appendix_converge}. We can find that, for CUB, the training converges around 16 epochs with the entropy guidance compared with 44 epochs without entropy guidance. Besides, the training accuracy  with entropy guidance is higher. On the contrary, the entropy slightly impairs the performance on SUN, which is consistent with our observation in \textit{Section Experiment}. This may be due to the lower average entropy of SUN.

\subsection{Hyper-parameters}\label{app_hyper-para}
\begin{figure}[htp]
\centering 
\begin{subfigure}{0.21\textwidth}
\centering
\includegraphics[width=\textwidth]{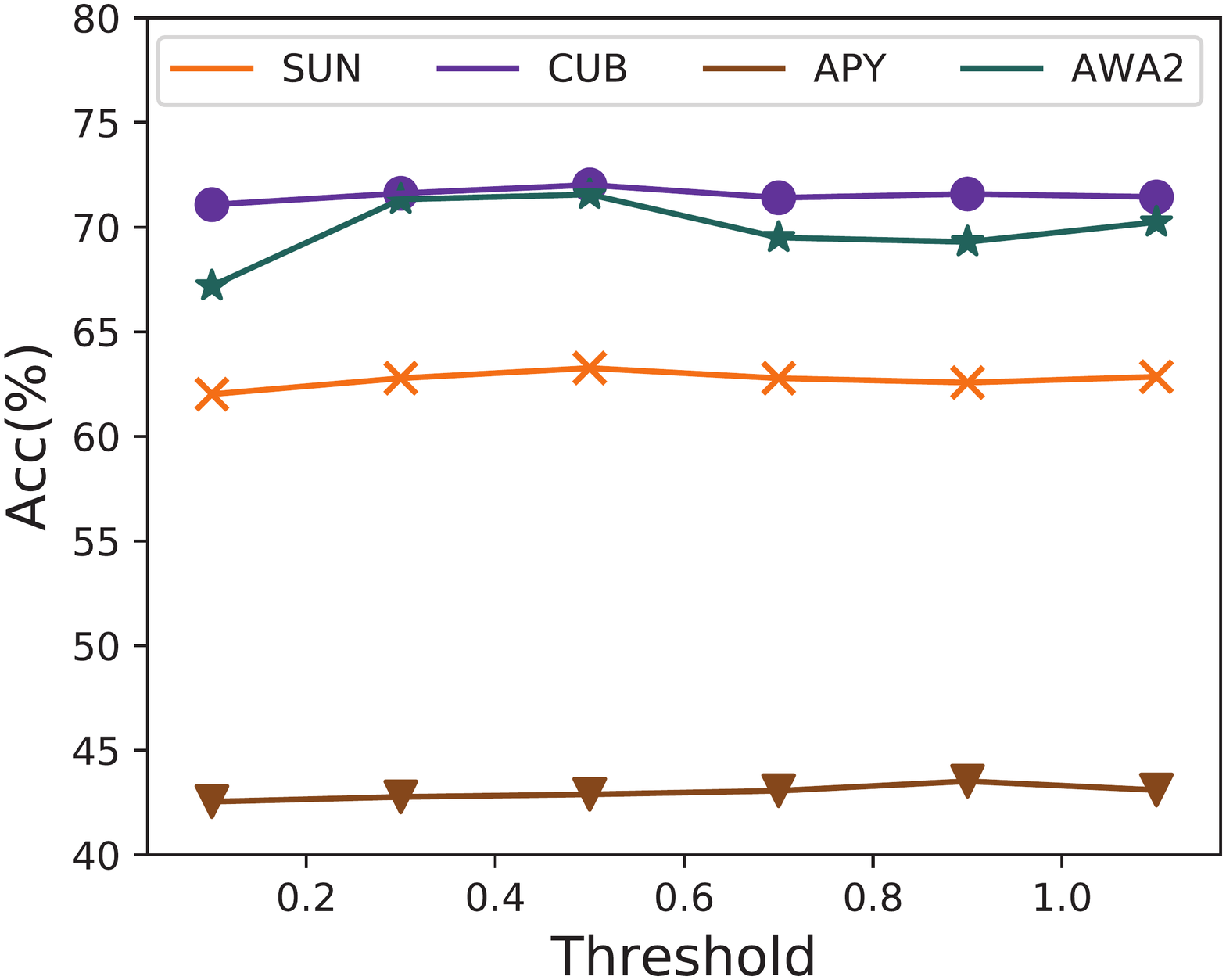}
\caption{Threshold analysis.}
\end{subfigure}\hfil 
\begin{subfigure}{0.21\textwidth}
\centering
\includegraphics[width=\textwidth]{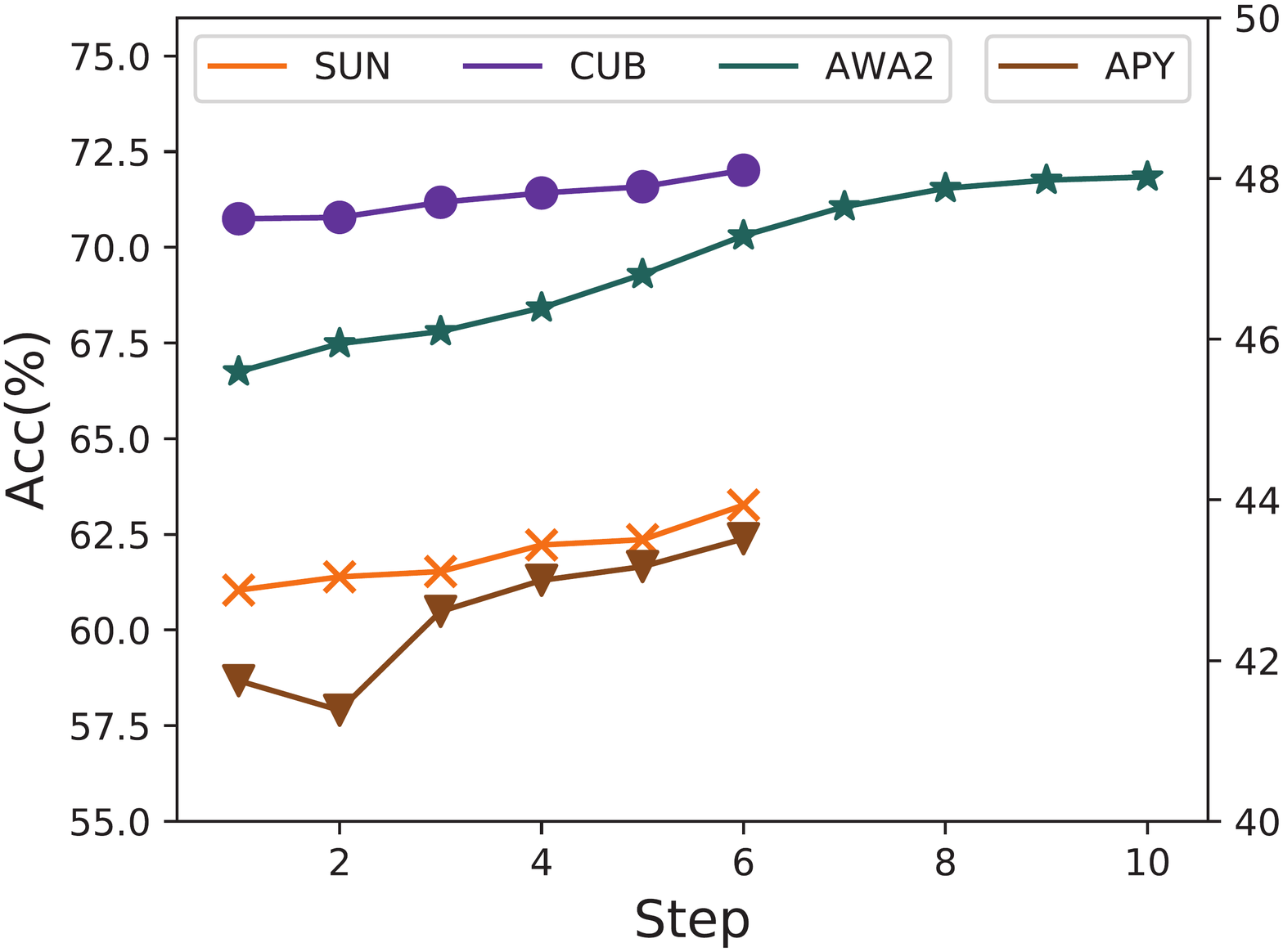}
\caption{Step-acc curve.}
\end{subfigure}\hfil 
    \caption{(a) Threshold hyper-parameter analysis and (b) step-acc curve on RPCNet.}
\label{appendix exp}
\end{figure}

\textbf{Threshold $\sigma$ of $\pi$}: For RPCNet, we show the results of average per-class accuracy of ZSL when varying $\sigma$ from 0.1 to 1.1 with a step of 0.2. The results in Figure~\ref{appendix exp}(a) are stable on $\sigma$ except for AwA2 dataset.

\textbf{Step-acc curve}: We fix the maximum step $T$ of RPCNet to be 6 for SUN, CUB, and aPY, whereas 10 for AwA2. The step-acc curves in Figure~\ref{appendix exp}(b) illustrate the changing tendency of accuracy when the locality is progressively explored. Overall, the accuracy increases as more steps are performed, and the improvements tend to be subtle after exploring sufficient localities.

\begin{figure}{r}
\centering 
\begin{subfigure}{0.3\textwidth}
\centerline{\includegraphics[width=\textwidth,height=6mm]{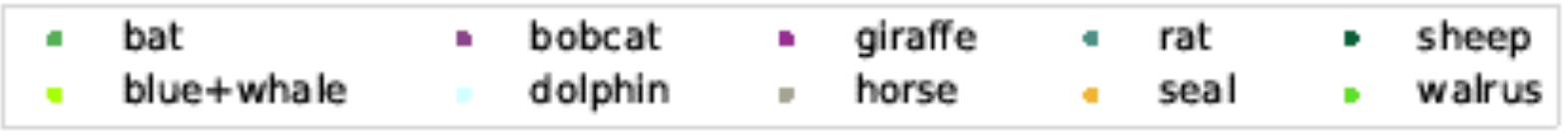}}
\end{subfigure}\\
\begin{subfigure}{0.2\textwidth}
\centering
\includegraphics[width=\textwidth]{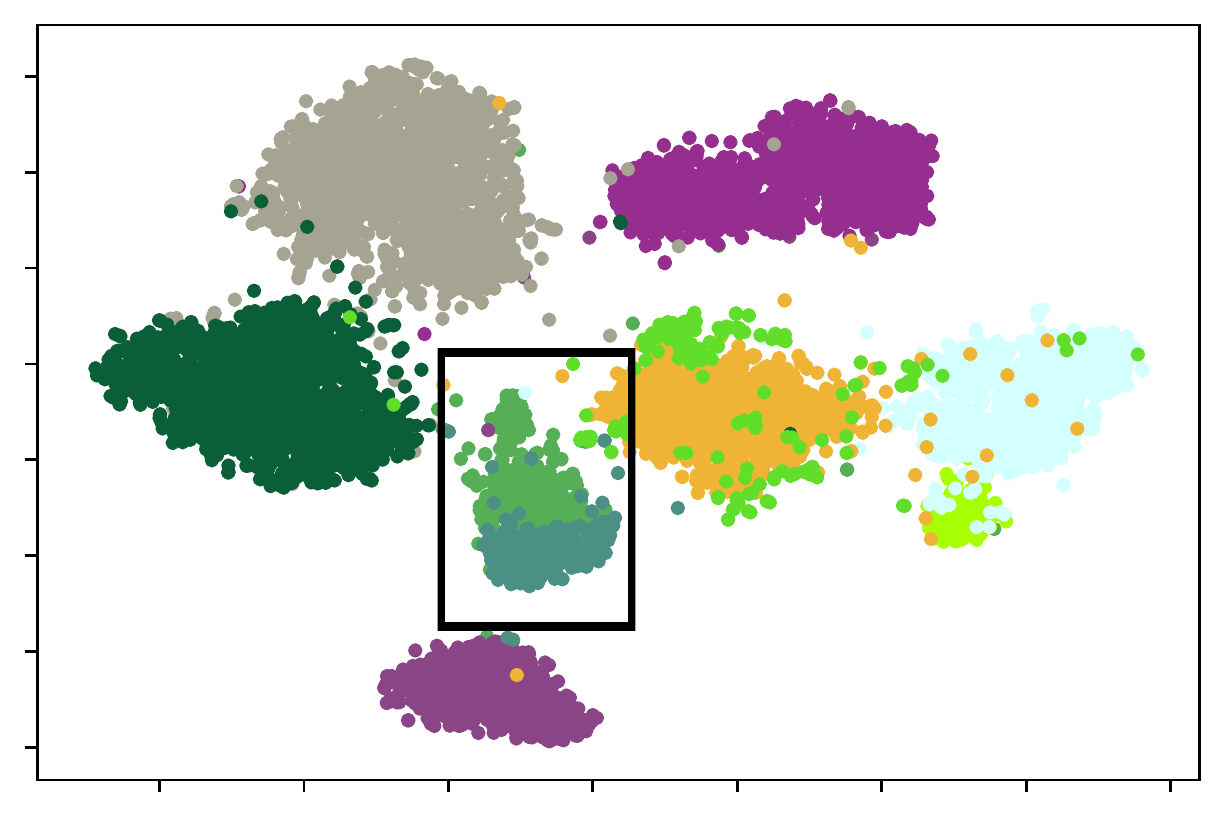}
\caption{Global embedding.}
\end{subfigure}\hfil 
\begin{subfigure}{0.2\textwidth}
\centering
\includegraphics[width=\textwidth]{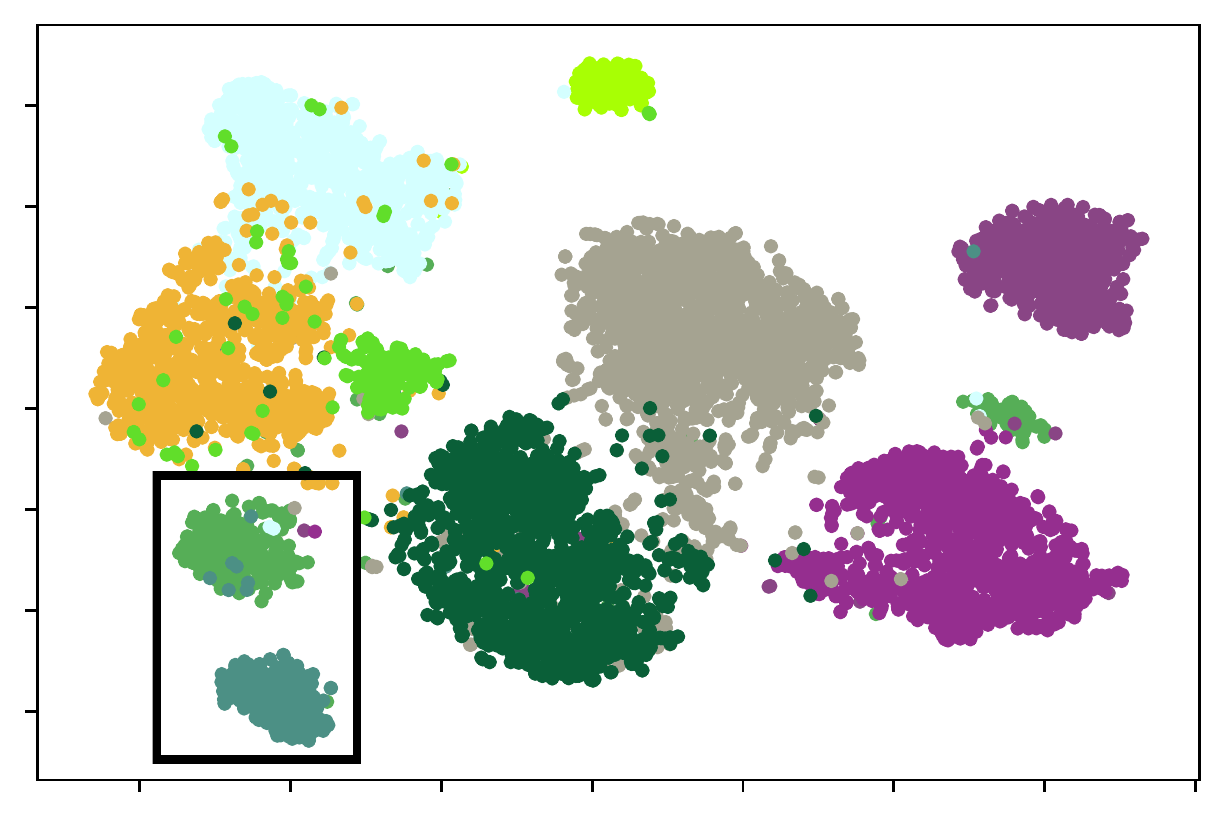}
\caption{Union embedding.}
\end{subfigure}\hfil 
    \caption{t-SNE visualization of unseen classes on AwA2. Rat and Bat are circled.}
\label{embdding}
\end{figure}

\subsection{Embedding Distribution Visualization}\label{app_embedding}
Figures~\ref{embdding} (a) and (b) visualize the distributions of the global embedding $\bar{e}_{g}^n$ and the union embedding $\bar{e}_u=\bar{e}_{g}^n+ {e}_{j}$ of unseen classes, respectively, on AwA2 by t-SNE visualization~\cite{van2008visualizing}. $e_{j}$ is the intermediate output of $f_{cj}$
The results show that
the global embedding can distinguish most classes but can still be confused on some unseen classes, such as \emph{bat} and \emph{rat} (circled in Figure~\ref{embdding}); in contrast, the union embedding, combined with localities, is discriminative enough to distinguish the confused classes.

\subsection{Progressive Selection Visualization}\label{app_selection}
\begin{figure}[!htb]
\centering
\includegraphics[width=\linewidth]{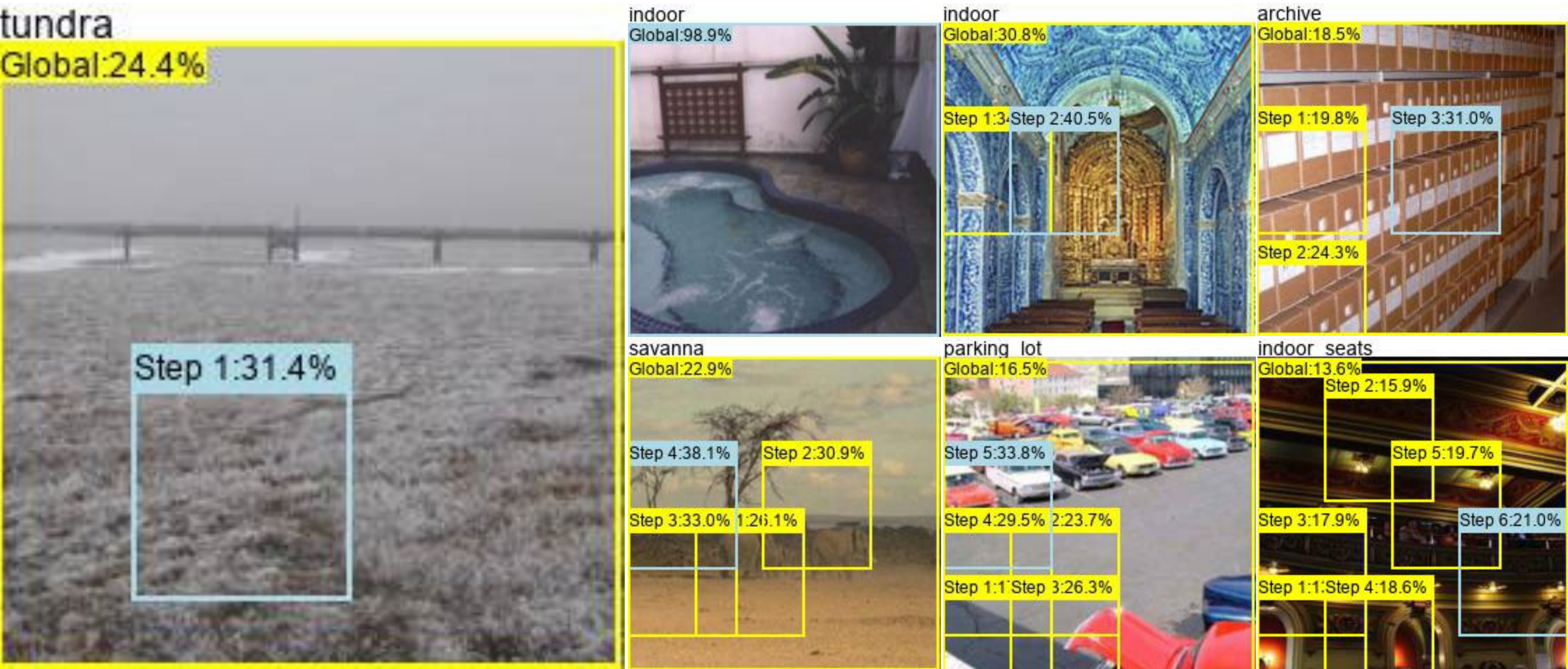}
\caption{Visualization of progressive locality selection on SUN. The labels above the boxes denote the step index and the prediction confidence after this selection. The box color indicates the prediction correctness at the current step (blue: correct; yellow: wrong).}
\label{sun_locality_visualization}
\end{figure} 
As shown in Fig~\ref{sun_locality_visualization}, different from the CUB dataset, the model tends to progressively choose representative regions for diverse objects on the SUN dataset.
For example, the proposed ERPCNet first focuses on the wall, 
then 
concentrates on decorations and chairs to distinguish indoor and indoor seats. This indicates that our method can progressively pick up the best locality to distinguish similar or diverse objects effectively. 

\subsection{Failure Modes of RL Component}\label{app_fail}
We performed a detailed investigation to explore failure modes of RL, including the definition, the statistics, and the reasons for failure mode.

We use trained models and do experiments on the CUB dataset in the ZSL setting. There are 2,697 pictures in the test set, with an accuracy rate of 72.5\%, i.e., 816 pictures are misclassified. There are two types of misclassifications: 1) The model keeps misclassifying the images during the entire decision-making process of extracting global information and exploring the localities; 2) The model first classifies the images correctly but then misclassifies the images after performing several steps of locality exploration.

We believe that the first kind of misclassification is because the images are beyond the classification capabilities of our model, and consider the second kind as the failure mode of the RL component. Specifically, 759 images belong to the first misclassification, and 57 images belong to the second misclassification, i.e., the failure mode. In most cases, RL components are qualified (57 failures compared with 1,938 images that are within the model capability). 

\begin{table}
	\centering
	\small
\begin{tabular}{cccccc}
\hline
Step 1  & Step 2  & Step 3  & Step 4 & Step 5 & Step 6  \\ 
\hline
\hline
8&3&13&3&17&13\\
\hline
\end{tabular} 
\caption{Failure occur steps.}
\label{table_failure}
\end{table}

\begin{table}
	\centering
	\small
\begin{tabular}{ccccccc}
\hline
Global & Step 1  & Step 2  & Step 3  & Step 4 & Step 5 & Step 6  \\ 
\hline
\hline
48.7&47.5&46.3&45.2&44.2&42.8&37.6\\
\hline
\end{tabular} 
\caption{Accuracy at different steps.}
\label{table_failure_acc}
\end{table}

We show the steps of failure occurrence and the number of failures in Table~\ref{table_failure}. We find that failure may occur after any step of locality exploration, and there is no obvious regularity in the number distribution.
We also find that in failure mode, no matter which steps the failure happens, the predicted probabilities of correct labels decrease as more localities are explored. Taking a failure image belonging to \textit{northern fulmar} (a seabird) as an example, the failure happens after exploring 5 localities. We show the trend of the predicted probability (\%) of the correct label in Table~\ref{table_failure_acc}. We can find that the probability declines throughout the process.

We observe the 57 images belonging to the failure mode and find that in these pictures, the main objects account for a relatively small area and are often hidden in cluttered environments, e.g., a bird hiding in a dense tree. Considering that the predicted probabilities of the correct labels decline during the decision-making process, we infer that the failure mode happens because: the initial prediction probability does not reach the threshold of RL, so the model continues to perform locality exploration; then the messy background is incorporated as localities, which introduces noise, making the probability of correct label decrease and finally leading to wrong results.
\end{document}